\documentclass{article} %
\usepackage[final]{colm2026_conference}

\usepackage{microtype}
\usepackage{hyperref}
\usepackage{url}
\usepackage{booktabs}
\usepackage{graphicx} %
\usepackage{enumitem}
\usepackage{multirow}
\usepackage{lineno}
\usepackage{amsmath}
\usepackage{stfloats}
\usepackage{caption}
\usepackage{subcaption}
\usepackage{xcolor}
\usepackage[table]{xcolor}
\usepackage{placeins}

\usepackage{xcolor}
\usepackage{changepage}

\definecolor{darkblue}{rgb}{0, 0, 0.5}
\hypersetup{colorlinks=true, citecolor=darkblue, linkcolor=darkblue, urlcolor=darkblue}

\title{Differentiable Faithfulness Alignment \\ for Cross-Model Circuit Transfer}

\author{Shun Shao \\
University of Cambridge \\
\texttt{ss3047@cam.ac.uk}
\And
Binxu Wang \\
Kempner Institute, Harvard University \\
\texttt{binxu\_wang@g.harvard.edu}
\AND
\begin{minipage}{0.503\textwidth}
Shay B. Cohen \\
\normalfont
University of Edinburgh \\
\texttt{scohen@inf.ed.ac.uk}
\end{minipage}
\And
\begin{minipage}{0.5\textwidth}
Anna Korhonen \\
\normalfont
University of Cambridge \\
\texttt{alk@cam.ac.uk}
\end{minipage}
\medskip
\AND
\begin{minipage}{0.5\textwidth}
Yonatan Belinkov \\
\normalfont
Kempner Institute, Harvard University \\
Technion \\
\texttt{belinkov@technion.ac.il}
\end{minipage}
}

\begin{document}

\ifcolmsubmission
\linenumbers
\fi

\maketitle

\newcommand{\shaycomment}[1]{\textcolor{blue}{#1 -- Shay}}

\begin{abstract}
Mechanistic interpretability has made it possible to localize circuits underlying specific behaviors in language models, but existing methods are expensive, model-specific, and difficult to scale to larger architectures. We introduce \textbf{Differentiable Faithfulness Alignment (DFA)}, a framework that transfers circuit information from a smaller source model to a larger target model through a learned differentiable alignment. DFA projects source-model node importance scores into the target model and trains this mapping with a soft faithfulness objective, avoiding full circuit discovery on the target model. We evaluate DFA on Llama-3 and Qwen-2.5 across six tasks spanning factual retrieval, multiple-choice reasoning, and arithmetic. The strongest results occur on Llama-3 $1$B$\rightarrow3$B, where aligned circuits are often competitive with direct node attribution and zero-shot transfer remains effective. Recovery weakens for larger source--target gaps and is substantially lower on Qwen-2.5, suggesting that transfer becomes harder as architectural and scaling differences increase. Overall, DFA consistently outperforms simple baselines and, in some settings, recovers target-model circuits with faithfulness comparable to or stronger than direct attribution. These results suggest that smaller models can provide useful mechanistic priors for larger ones, while highlighting both the promise and the limits of node-level cross-model circuit alignment.\footnote{Code is available at \url{https://github.com/jasonshaoshun/dfa-circuits}.}
\end{abstract}

\section{Introduction}

Recent mechanistic interpretability work has shown that LLM behaviors can often be reverse-engineered through circuit analysis \citep{wang2022interpretability, hanna2024eapig, marks2024saecircuits}. A central question is whether these findings generalize across model sizes, tasks, architectures, and training regimes \citep{wang2024universality}. Although the field lacks a unified theory of such generalization \citep{trott2025theory}, growing empirical evidence suggests that mechanisms are often at least partially shared: algorithms can persist across scale and training \citep{lieberum2023scale, tigges2024llm, prakash2023fine}, related tasks can reuse sub-circuits \citep{merullo2024circuitreuse, lan2024towards, mondorf2025modular}, and mechanistic similarity can remain even when exact structural correspondence is weak \citep{wang2024universality, nikankin2025sametask}. This question is scientifically important and practically urgent: existing circuit-localization methods are computationally expensive and become increasingly difficult to apply reliably as model size grows \citep{hanna2024eapig}. If larger models are often not solving tasks with entirely different mechanisms from smaller ones, then circuits discovered in smaller models may provide useful signal for predicting target-model circuits without rerunning costly localization procedures from scratch.

\begin{figure}[!t]
  \centering
  \includegraphics[width=\linewidth]{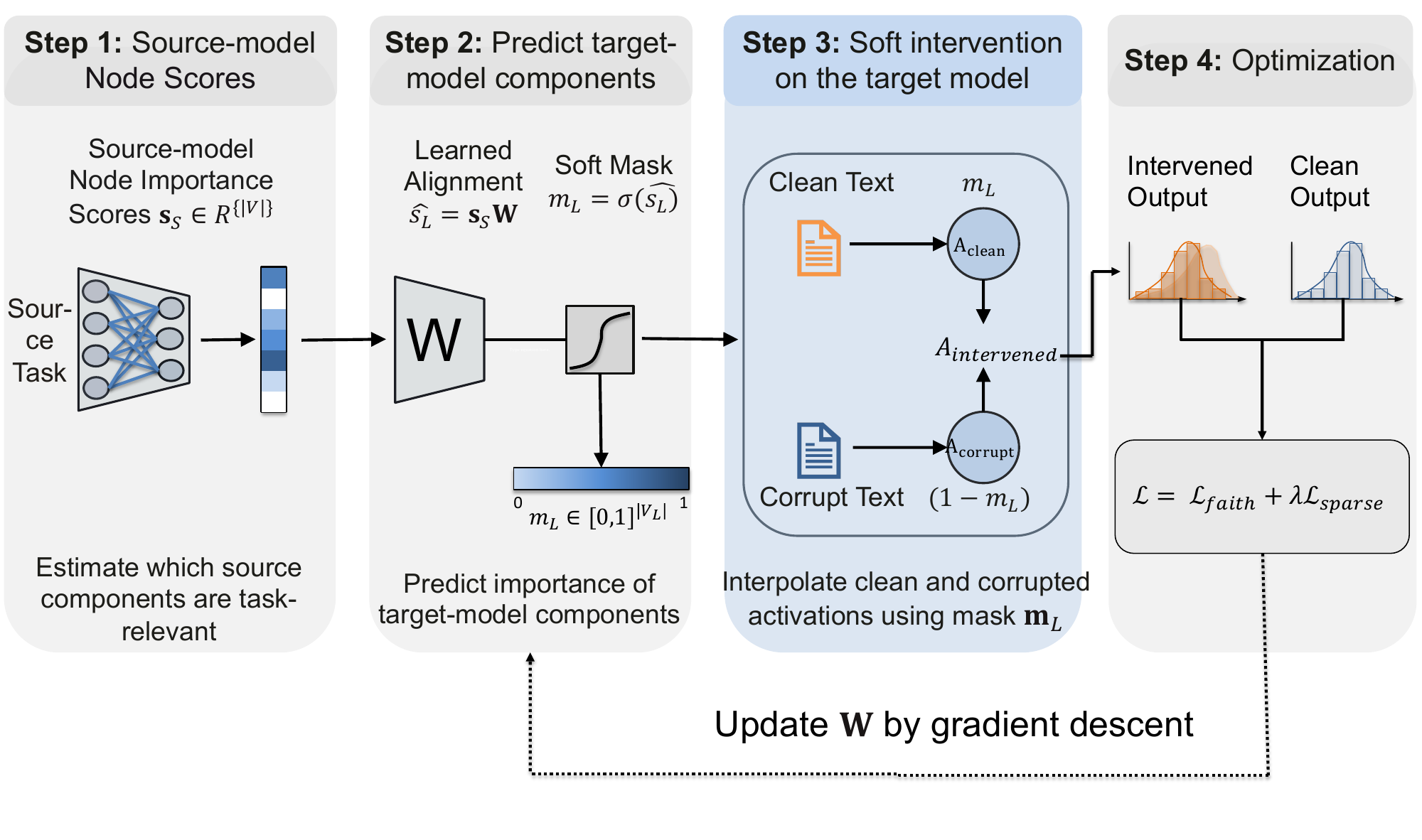}
\caption{Overview of \textbf{Differentiable Faithfulness Alignment (DFA)}. Source-model node importance scores are projected into the target model through a learned alignment matrix $W$ and converted into a soft mask $m_L$. This mask interpolates between clean and corrupted target-model activations, producing an intervened output distribution. The alignment matrix is then optimized with a faithfulness loss and an $L_1$ sparsity penalty, allowing source-model circuit information to predict target-model circuits without running full circuit localization on the target model.}

\label{fig:dfa-overview}
\end{figure}

To this end, we propose \textbf{Differentiable Faithfulness Alignment (DFA)}, a framework for transferring circuit information across models of different scales. Rather than localizing circuits independently in both models and comparing them post hoc, DFA treats cross-model circuit transfer as a predictive alignment problem. Starting from importance scores for circuit nodes in a smaller, more transparent \emph{source} model, DFA learns a correspondence to components in a larger \emph{target} model. This learned correspondence is then used to predict a soft target-side circuit, indicating which target components are likely to be causally important for the task or behavior of interest. The correspondence is learned by optimizing a differentiable faithfulness objective on the target model, allowing DFA to recover target-side circuits without requiring a gold target circuit during training. Figure~\ref{fig:dfa-overview} provides an overview of the full pipeline.

Our results over two model families and six tasks show that circuit information can partially transfer across model scales: source-model circuits can often recover faithful target-model circuits without direct attribution on the target model. In all cases, DFA outperforms simple baselines, showing that its gains depend on a learned source--target correspondence rather than trivial architectural priors. Transfer is strongest for smaller within-family gaps, especially Llama-3 $1$B$\rightarrow3$B, and weakens for larger gaps and for Qwen-2.5. Strikingly, DFA can sometimes produce target-model circuits that are more faithful than direct node attribution on the target model, suggesting that smaller models may provide cleaner mechanistic signals for alignment. At the same time, the decline across harder model pairs highlights clear limits of node-level alignment. In summary, our main contributions are:
\begin{itemize}[leftmargin=*,nosep]
    \item \textbf{Differentiable Alignment Framework:} We introduce \textbf{Differentiable Faithfulness Alignment (DFA)}, a method that learns to map source-model circuit scores to target-model circuits through a differentiable faithfulness objective.
    \item \textbf{Cross-Model Circuit Recovery:} We show that source-model circuits can recover faithful target-model circuits across multiple model pairs and transfer regimes, including zero-shot transfer.
    \item \textbf{Evidence for Learned Correspondence:} Through ablations, we show that DFA's gains depend on structured source-model signal and a learned source--target mapping, rather than on random masks or simple architectural priors.
    \item \textbf{Limits of Transferability:} We identify clear limits of node-level alignment, with performance degrading as the source--target gap increases.
\end{itemize}

\section{Related work}
\label{sec:related-work}

Understanding whether mechanistic findings generalize across tasks, scale, and model families is a central challenge for interpretability \citep{wang2024universality, trott2025theory}. This question is especially important for large language models, where direct circuit analysis remains computationally expensive and often partial. Direct circuit discovery at scale is still relatively rare, with most work focusing on models around 8B parameters or smaller. Notable exceptions nevertheless suggest substantial continuity of mechanism across model variants. For example, \citet{lieberum2023scale} recover mechanisms in Chinchilla 70B, while \citet{tigges2024llm} show that similar algorithms recur across training checkpoints and model sizes despite variation in the specific components involved. Likewise, fine-tuning often strengthens existing mechanisms rather than replacing them with entirely new ones \citep{prakash2023fine}. Together, these works suggest that scaled or adapted models often do not solve tasks using entirely different underlying algorithms. Circuits are also functionally reused across related tasks \citep{merullo2024circuitreuse, lan2024towards, mondorf2025modular}. However, structural overlap is neither guaranteed nor strictly necessary for functional similarity. For instance, circuits for identical tasks can differ structurally in multimodal settings \citep{nikankin2025sametask}, and higher overlap does not always yield higher faithfulness \citep{hanna2024eapig}. Mechanistic similarity even spans distinct architectures such as Transformers and Mamba \citep{wang2024universality}. These findings encourage cross-model comparisons, but they also imply that requiring exact component or architectural matching is too restrictive for studying transferability. Conceptually, our approach resembles Distributed Alignment Search  \citep[DAS;][]{geiger2024finding}, which learns correspondences between causal variables and neural representations. However, rather than aligning a high-level causal model to a single network, we explicitly align circuits between two different language models, moving the comparison from structural overlap to learned functional alignment. %
Other work used linear mappings to align two models, differing in finetuned data (general versus specific) rather than size to identify behavior in activation space \citep{zhao-etal-2023-joint,zhao-etal-2024-layer}. %

\section{Methodology}

We formalize circuit discovery in the computational-graph view of Transformers and then define the problem of cross-model circuit alignment. Our goal is to recover a target-model circuit from a source-model circuit, without running a full circuit-localization procedure on the target model. We introduce \textbf{Differentiable Faithfulness Alignment (DFA)}, which learns a functional mapping between model components by optimizing a differentiable faithfulness objective. Figure~\ref{fig:dfa-overview} gives an overview of the full DFA pipeline, from source-model circuit scores to soft intervention and optimization on the target model.

\subsection{Problem setup: the circuits framework}

We assume a computational graph $G=(V,E)$ is given for a transformer model $\mathcal{M}$ (\S\ref{section:setup}). Nodes $V$ correspond to model components such as attention heads or MLP modules, and edges represent residual-stream dependencies between them. An edge $(u,v)\in E$ indicates that the output of component $u$ contributes to the input of component $v$. The graph therefore captures how information propagates from input embeddings to final logits.

A circuit is a task-relevant subgraph of this computation \citep{elhage2021mathematical}. We say that a circuit is \emph{faithful} if preserving the computation inside the circuit is sufficient to recover the model's behavior on the task \citep{wang2022interpretability,hanna2024eapig}, and the components are sufficient to restore the model's task-relevant output. Following MIB \citep{mib-2025}, the faithfulness of a candidate circuit $C$ relative to the full model $\mathcal{M}$ is defined as $ f(C,\mathcal{M};m)=\frac{m(C)-m(\emptyset)}{m(\mathcal{M})-m(\emptyset)}$, where $m$ is a task-specific performance metric, $m(C)$ is the performance when retaining only the circuit, $m(\mathcal{M})$ is the clean-model performance, and $m(\emptyset)$ is the fully corrupted baseline. %

In this work, we focus on node-level circuit localization \citep{syed-etal-2024-attribution}. For a model with node set $V$, we represent a circuit as a node-importance score \emph{row vector} $\smash{\mathbf{s}\in\mathbb{R}^{|V|}}$, where each entry measures the importance of one component to the behavior of interest. Thresholding this score vector yields circuits of different sizes, whose faithfulness can then be evaluated.\footnote{At the node level, thresholding importance scores need not produce a connected subgraph in the strict graph-theoretic sense. As in prior node-level circuit work, we use ``circuit'' operationally to denote the selected set of task-relevant components, and evaluate it by faithfulness rather than by an explicit connectivity constraint.}

\paragraph{The alignment problem.}

Let $\mathcal{M}_S$ and $\mathcal{M}_L$ be the source and target models, with node sets $V_S$ and $V_L$, respectively. Given a source-model circuit score vector $\smash{\mathbf{s}_S\in\mathbb{R}^{|V_S|}}$, our goal is to predict a corresponding target-model circuit without running direct attribution on $\mathcal{M}_L$. Treating score vectors as row vectors, we learn an alignment matrix $\smash{W \in \mathbb{R}^{|V_S|\times |V_L|}}$ that projects source-model scores into the target model: $\hat{\mathbf{s}}_L = \mathbf{s}_S W.$
The projected vector $\hat{\mathbf{s}}_L\in\mathbb{R}^{|V_L|}$ defines a target-side score profile over components. At test time, we treat $\hat{\mathbf{s}}_L$ as the predicted target-model circuit and evaluate it in the same way as a directly attributed circuit. The central question is therefore whether a source-model circuit, together with a learned alignment matrix $W$, is sufficient to recover a faithful circuit in the target model.

\subsection{Differentiable faithfulness alignment}

The key innovation in DFA is a differentiable faithfulness intervention that allows us to learn the alignment matrix $W$ without access to gold target circuits. Source-model importance scores are projected into the target model and converted into a soft mask over target components. This mask controls a soft interpolation between clean and corrupted activations in the target model, and $W$ is optimized so that the intervened target model recovers the clean behavior. In this way, DFA learns a causally faithful soft target-side circuit without running direct attribution on the target model.

\subsubsection{Soft target-model circuit}

Given the projected score vector $\hat{\mathbf{s}}_L = \mathbf{s}_S W$, we convert it into a continuous mask over target-model components: $\mathbf{m}_L = \sigma(\hat{\mathbf{s}}_L)$
where $\sigma(\cdot)$ denotes the \emph{elementwise} sigmoid function. We use an elementwise sigmoid rather than a softmax because the target circuit is not a probability distribution over components: multiple target nodes may be important simultaneously, and each component should therefore be gated independently. Each entry $\smash{\mathbf{m}_L^{(j)}\in[0,1]}$ specifies how strongly target component $j$ is preserved. Components with larger mask values are pushed more strongly toward their clean activations, while components with smaller mask values remain closer to their corrupted activations.

Although $\mathbf{m}_L$ is computed from a single source score vector at inference time, $W$ is not learned for one task in isolation. Instead, it is optimized across many source-score vectors and training tasks, so that it captures a reusable source--target correspondence rather than a single task-specific mask.

\subsubsection{Differentiable faithfulness intervention}

To test whether a predicted target-side circuit is sufficient for the behavior of interest, we use the standard clean/corrupted intervention setting from circuit analysis. For each clean input $x$, we define a matched corrupted input $x'$ that preserves the overall prompt structure but alters the task-relevant signal, yielding a controlled counterfactual run. Activations from this corrupted run therefore provide a natural reference for what the model computes when the relevant information is disrupted.

Let $A_{\mathrm{clean}}^{(j)}$ denote the activation of target component $j$ cached from the clean run. Let $A_{\mathrm{corrupt}}^{(j)}$ denote its activation during the forward pass on the corrupted prompt, after any upstream interventions and immediately before applying its own mask. We define the intervened activation as

\begin{equation}
\tilde{A}^{(j)} = \mathbf{m}_L^{(j)} A_{\mathrm{clean}}^{(j)} + \bigl(1-\mathbf{m}_L^{(j)}\bigr) A_{\mathrm{corrupt}}^{(j)}.
\end{equation}
Components with larger mask values are restored more strongly from the clean run, while components with smaller mask values remain closer to their corrupted activations. This soft interpolation makes the faithfulness objective differentiable with respect to $W$, allowing DFA to learn target-side circuit structure without access to gold target circuits.

\subsubsection{Training objective}

We train $W$ by minimizing a loss function $\mathcal{L}$ composed of a faithfulness term and a sparsity penalty.

\paragraph{Faithfulness Loss.}
We measure the divergence between the output distribution of the clean large model, $P_{\text{clean}}$, and the output distribution of the model under the differentiable intervention, $P_{\text{intervened}}$. We use the Kullback-Leibler (KL) divergence:

\begin{equation}
    \mathcal{L}_{\text{faith}} = D_{KL}(P_{\text{clean}} \parallel P_{\text{intervened}}).
\end{equation}

Minimizing this term encourages $W$ to assign high importance scores to the subset of large model components that are causally necessary to restore the correct output distribution.

\paragraph{Sparsity Penalty.}
To prevent the trivial solution where $\mathbf{m}_L \approx \mathbf{1}$ (which would simply select the entire model and trivially recover performance), we impose an $L_1$ penalty on the mask to encourage sparsity:

\begin{equation}
    \mathcal{L}_{\text{sparse}} = \frac{1}{|V_L|} \sum_{j=1}^{|V_L|} m_L^{(j)}.
\end{equation}

Here, $j$ indexes target-model components (i.e., nodes in $V_L$), and $m_L^{(j)} \in [0,1]$ is the soft gate for component $j$. The second term therefore penalizes the average mask mass, encouraging sparse target-side circuits while avoiding the trivial solution that preserves all target components. The final objective is:

\begin{equation}
    \mathcal{L} = \mathcal{L}_{\text{faith}} + \lambda \mathcal{L}_{\text{sparse}}.
\end{equation}

By optimizing this objective across a diverse set of training tasks, $W$ learns a mapping of functional roles from the small model architecture to the large model architecture. The computational benefit is amortized across tasks rather than one-shot; see Appendix~\ref{app:compute}.

\section{Experiments}
\label{sec:experiments}

\begin{figure}[t]
  \centering
  \includegraphics[width=\linewidth]{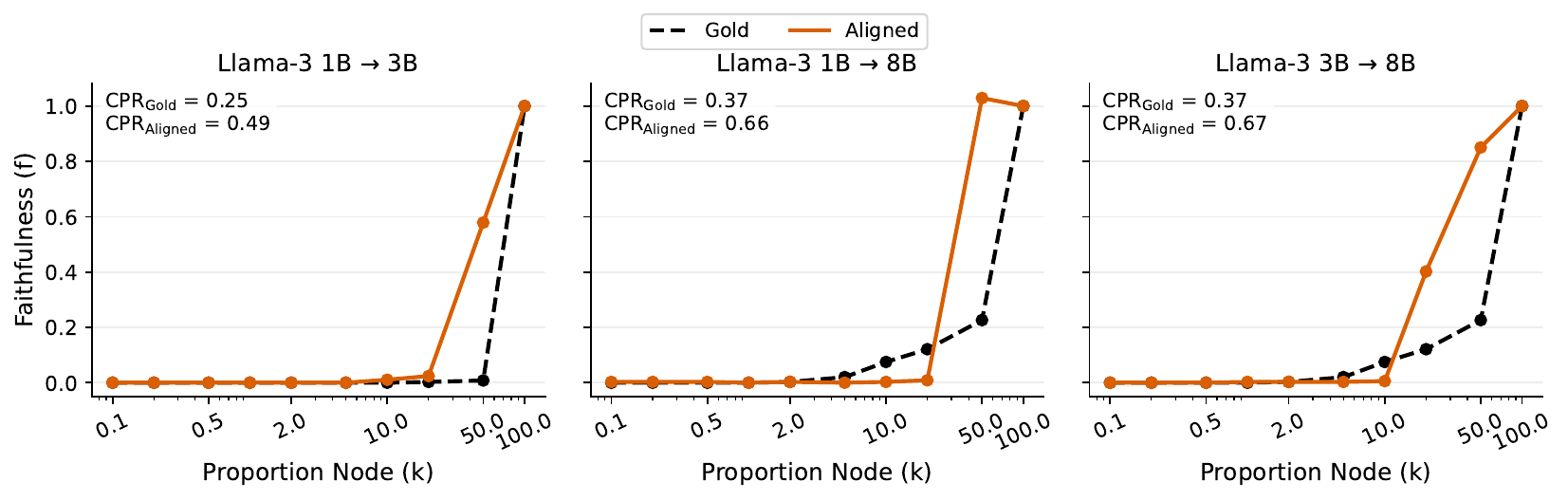}

\caption{Direct target-model circuits (\textbf{Gold}) and DFA-predicted circuits (\textbf{Aligned}) on \textsc{Arithmetic-Subtraction} for three Llama-3 source--target pairs. The x-axis shows the proportion of retained nodes $k$, and the y-axis shows faithfulness $f$. Aligned circuits closely track the gold faithfulness curves and often achieve higher CPR.}

  \label{fig:1b-eap-ig-inputs-arith-sub-llama-best}
\end{figure}

We test whether node-level circuit scores from smaller language models can be mapped into larger models through a learned differentiable alignment, and whether the resulting target-model circuits remain faithful under task transfer, model scaling, and architectural variation.
\textbf{(1)} Can DFA recover faithful target-model circuits without full circuit discovery on the target model? (\S\ref{subsec:exp_main})
\textbf{(2)} Do gains arise from a meaningful cross-model correspondence rather than a generic sparse mask or static architectural prior? (\S\ref{subsec:exp_ablations})
\textbf{(3)} Does the approach remain effective as the source--target gap grows and across model families? (\S\ref{subsec:exp_scaling})
\textbf{(4)} Does the learned alignment capture task-stable structure that transfers beyond the training task? (\S\ref{subsec:exp_heatmap})

\subsection{Experimental setup}
\label{subsec:exp_setup}
\label{section:setup}

\paragraph{Models and tasks.}

We evaluate on Llama-3 and Qwen-2.5. Our main results focus on Llama-3 $1$B$\rightarrow3$B, with additional model-pair analyses reported in the appendix for larger Llama-3 and Qwen-2.5 models. We follow the task setup of MIB \citep{mib-2025}and evaluate on the same six tasks: \textsc{IOI}, \textsc{MCQA}, \textsc{Arithmetic Addition}, \textsc{Arithmetic Subtraction}, \textsc{ARC-Easy}, and \textsc{ARC-Challenge}. 

\paragraph{Node-level source circuits.}

For each source model, we compute node-level importance scores using \textsc{Node Attribution Patching} (\textsc{NAP}; \citealt{nanda2023attribution, syed-etal-2024-attribution}), \textsc{Node Attribution Patching with Integrated Gradients--Activations} (\textsc{NAP-IG-Activations}), and \textsc{Node Attribution Patching with Integrated Gradients--Inputs} (\textsc{NAP-IG-Inputs}) by \cite{hanna2024eapig, marks2024saecircuits}. We also run these same attribution methods directly on the target model as evaluation references. DFA then maps the source-model scores into the target model through a learned alignment, yielding predicted target-model importance scores.

At node level, we include only attention heads and MLP blocks in the circuit representation. This yields source--target node-set sizes $(|V_S|,|V_L|)=(528,700)$, $(528,1056)$, and $(700,1056)$ for the three Llama-3 settings, and $(360,364)$, $(360,612)$, and $(364,612)$ for the three Qwen2.5 settings.

\paragraph{Evaluation metric.}
We evaluate predicted target-model circuits using \textbf{CPR} (Circuit Performance Ratio), following \citet{mib-2025}. Given a ranking of target-model components induced by a circuit score vector, we evaluate faithfulness $f$ across circuit sizes by retaining the top-$k$ proportion of components and measuring behavioral recovery under intervention. CPR is then the area under this faithfulness-versus-circuit-size curve. Higher CPR therefore indicates that a method recovers faithful behavior consistently across circuit sizes.

\paragraph{Transfer settings.}
We evaluate DFA under three transfer regimes.

\begin{itemize}[leftmargin=*,nosep]
    \item \textbf{In-Distribution:} We train $W$ and evaluate it on the same task (with different training/test splits). This measures whether the alignment has sufficient capacity to fit a task-specific source--target correspondence.
    \item \textbf{Near-Distribution:} We train on a closely related task and test on its paired counterpart: \textsc{Arithmetic Addition} $\leftrightarrow$ \textsc{Arithmetic Subtraction}, and \textsc{ARC-Easy} $\leftrightarrow$ \textsc{ARC-Challenge}. This probes transfer within a shared task family.
    \item \textbf{Far-Transfer / Zero-Shot:} For a target task $t$, we train $W$ on all remaining tasks and evaluate on $t$. This leave-one-task-out setting tests whether the learned alignment functions as a task-general translator rather than a task-specific fit.
\end{itemize}

These regimes are not matched-budget comparisons; training sizes are reported in Appendix ~\ref{app:transfer-sizes}.

\subsection{Main result: recovering target-model circuits from small-model signals}
\label{subsec:exp_main}

\begin{table*}[!t]
\centering
\small
\setlength{\tabcolsep}{3pt}
\renewcommand{\arraystretch}{1.05}
\begin{tabular}{l p{3.6cm} ccccccc}
\toprule
\textbf{Method} & \textbf{Transfer Setting} & \textbf{IOI} & \textbf{MCQA} & \textbf{Arith +} & \textbf{Arith -} & \textbf{ARC-E} & \textbf{ARC-C} & \textbf{Avg.} \\
\midrule
Baseline & Random Alignment & 0.25 & 0.25 & 0.25 & 0.25 & 0.25 & 0.25 & 0.25 \\
\midrule
\multirow{4}{*}{NAP-IG-Inputs} & Gold (Upper Bound) & 0.28 & \underline{0.47} & 0.26 & 0.25 & \underline{0.52} & \textbf{0.48} & 0.38 \\
& DFA (In-Distribution) & \textbf{0.35} & 0.41 & 0.29 & 0.45 & 0.41 & \underline{0.39} & 0.38 \\
& DFA (Near-Distribution) & - & - & \underline{0.43} & \underline{0.47} & \textbf{0.54} & 0.37 & \textbf{0.45} \\
& DFA (Far/Zero-Shot) & \underline{0.31} & \textbf{0.48} & \textbf{0.51} & \textbf{0.49} & 0.47 & 0.38 & \underline{0.44} \\

% \midrule
% \multirow{3}{*}{NAP-IG-Inputs}& Non-linear DFA (In-Distribution) & \textbf{0.34} & \textbf{0.50} & \underline{0.53} & \underline{0.53} & \textbf{0.57} & \underline{0.52} & \underline{0.50} \\
% & Non-linear DFA (Near-Distribution) & - & - & \textbf{0.54} & \textbf{0.55} & 0.51 & \textbf{0.57} & \textbf{0.54} \\
% & Non-linear DFA (Far/Zero-Shot) & \underline{0.33} & 0.27 & 0.40 & 0.40 & 0.27 & 0.27 & 0.32 \\

% \midrule
% \multirow{3}{*}{NAP-IG-Inputs}&  Supervised Regression (In-Distribution) & 0.25 & 0.25 & 0.26 & 0.25 & 0.43 & 0.40 & 0.31 \\
% & Supervised Regression (Near-Distribution) & - & - & 0.25 & 0.25 & 0.42 & 0.41 & 0.33 \\
% & Supervised Regression (Far/Zero-Shot) & 0.25 & \underline{0.43} & 0.25 & 0.25 & 0.25 & 0.41 & 0.31 \\

\bottomrule

\end{tabular}
\caption{Circuit Recovery Performance (Llama-3 (1B $\to$ 3B)). We report faithfulness using CPR. Near-distribution evaluation is only applicable to Arithmetic and ARC task pairs.}
\label{tab:leaderboard-llama3-1b-to-llama3-3b}
\end{table*}

Table~\ref{tab:leaderboard-llama3-1b-to-llama3-3b} presents our main result on Llama-3 $1$B$\rightarrow3$B using \textsc{NAP-IG-Inputs}, with results for other model pairs and attribution methods deferred to Appendix~Tables~\ref{tab:leaderboard-llama3-1b-to-llama3-3b：eap-and-eap-ig-activations}--\ref{tab:leaderboard-qwen2-5-1-5b-to-qwen2-5-3b}. Aligned circuits substantially outperform the random baseline, and the zero-shot setting is often competitive with the in-distribution setting. This suggests that the learned matrix $W$ is not simply fitting a single task, but captures a source-to-target functional correspondence that transfers across tasks. Figure~\ref{fig:1b-eap-ig-inputs-arith-sub-llama-best} provides a representative curve-level comparison: across multiple source--target pairs, the aligned circuits recover faithfulness profiles similar to those of direct target attribution and often achieve comparable CPR. A striking pattern is that aligned circuits can in some settings match the faithfulness of direct target-model attribution, and occasionally exceed it. We do not interpret this as a contradiction. Rather, circuit localization methods scale poorly with model depth and width \citep{hanna2024eapig}: attribution can produce cleaner and more faithful circuits on smaller models than on larger ones. Our results therefore suggest that DFA can leverage stronger source-model signals and map them into the target model's node space, recovering target-model circuits with competitive faithfulness even when direct target attribution is less reliable.

Across these results, \textsc{NAP-IG-Inputs} provides the strongest transfer signal overall, while \textsc{NAP} and \textsc{NAP-IG-Activations} remain generally above the random baseline but are less consistent. Near-distribution transfer is often particularly strong for arithmetic and ARC, suggesting that related tasks induce similar source--target correspondences. At the same time, performance varies across model pairs, with transfer generally stronger for Llama-3 than for Qwen-2.5, indicating that cross-model transfer is meaningful but not uniform.

\subsection{Ablations: is the gain due to real alignment?}

\label{subsec:exp_ablations}

\begin{table*}[t]
\centering
\small
\setlength{\tabcolsep}{3pt}
\renewcommand{\arraystretch}{1.05}
\begin{tabular}{l >{\raggedright\arraybackslash}p{4.2cm} cccccc}
\toprule
\textbf{Method} & \textbf{Setting / Control} & \textbf{IOI} & \textbf{MCQA} & \textbf{Arith +} & \textbf{Arith -} & \textbf{ARC-E} & \textbf{ARC-C} \\
\midrule
\multirow{6}{*}{NAP-IG-Inputs} & 1. Random $W$ (Lower Bound) & 0.25 & 0.25 & 0.25 & 0.25 & 0.25 & 0.25 \\
& 2. Scrambled input $s$ & 0.28 & 0.27 & \underline{0.34} & 0.25 & 0.40 & 0.25 \\
& 3. Permuted $W$ columns & 0.27 & \underline{0.30} & 0.25 & \underline{0.25} & 0.28 & 0.26 \\
& 4. Heuristic Depth Mean & 0.25 & 0.25 & 0.25 & 0.25 & 0.25 & 0.25 \\
\addlinespace[4pt]
& \textbf{DFA (Zero-shot)} & \underline{0.31} & \textbf{0.48} & \textbf{0.51} & \textbf{0.49} & \underline{0.47} & \underline{0.38} \\
& \textbf{DFA (Best)} & \textbf{0.35} & \textbf{0.48} & \textbf{0.51} & \textbf{0.49} & \textbf{0.54} & \textbf{0.39} \\
\bottomrule
\end{tabular}

\caption{Validation study for target model (Llama-3 (1B $\to$ 3B)). The  \textbf{Alignment (Best)} reports the strongest result across zero-shot, near-distribution, and in-distribution transfer.}
\label{tab:ablations-summary-llama3-1b-to-llama3-3b}
\end{table*}

Table~\ref{tab:ablations-summary-llama3-1b-to-llama3-3b} tests whether DFA succeeds by learning a meaningful source--target correspondence, rather than because the task is easy or because target performance can be recovered without the true source-model signal. We compare alignment against the following controls:

\begin{itemize}[leftmargin=*,nosep]
    \item \textbf{Random $W$ (Lower Bound):} $W$ is initialized with the same scaling factor but not trained, testing whether any random sparse mask can succeed.
    
    \item \textbf{Scrambled $s$ (Signal Control):} $W$ is trained normally, but the source-model score vector $s$ is permuted at inference time. This preserves the marginal score distribution while destroying semantic assignment to source components.

    \item \textbf{Permuted $W$ (Topology Control):} $W$ is trained normally, but its columns are permuted at inference time. This preserves the global weight distribution while breaking the learned source--target correspondence, testing whether the topology of the learned mapping matters.

    \item \textbf{Heuristic $W$ (Architectural Prior Control):} $W$ is not trained, but fixed from architecture alone: each source layer is mapped to the two nearest target layers by relative depth with interpolation. This tests whether a simple architecture-based prior, loosely motivated by prior layer-similarity analyses, is already sufficient \citep{wu-etal-2020-similarity, pmlr-v97-kornblith19a}.
\end{itemize}

Both alignment conditions consistently outperform the controls. \textbf{Random $W$} remains uniformly weak, ruling out the possibility that an arbitrary untrained mask is sufficient. \textbf{Scrambling the source signal $s$} also hurts performance, indicating that DFA relies on which source-model components are important, not just on the learned matrix alone. \textbf{Permuting the columns of $W$} substantially degrades performance, showing that success depends on the learned source--target correspondence rather than merely on the global distribution of weights. Finally, \textbf{Heuristic $W$} performs poorly across tasks, showing that a simple structural prior based only on relative depth and component type is not enough to explain the gains. Taken together, these results support the claim that DFA works by combining structured source-model circuit information with a learned cross-model mapping. Additional diagnostic ablations in Appendix~\ref{app:diagnostic-ablations} show that supervised target-score regression underperforms DFA, while nonlinear alignment improves in-distribution and near-distribution transfer but hurts zero-shot transfer, suggesting an expressivity--generalization tradeoff.

\subsection{Scaling across model gaps and families}
\label{subsec:exp_scaling}

\begin{table*}[!t]
\centering

\resizebox{\textwidth}{!}{
\begin{tabular}{llcccc}
\toprule
\textbf{Method} & \textbf{Target Model Pair} & \textbf{Baseline} & \textbf{DFA} & \textbf{Gold} & \textbf{Faithfulness Recovery} \\
\midrule
\multirow{6}{*}{NAP-IG-Inputs} & Llama-3 (1B $\to$ 3B) & 0.25 & 0.48 & 0.47 & 103.3\% \\
& Llama-3 (1B $\to$ 8B) & 0.25 & 0.51 & 1.02 & 50.0\% \\
& Llama-3 (3B $\to$ 8B) & 0.25 & 0.46 & 1.02 & 44.7\% \\
& Qwen-2.5 (0.5B $\to$ 1.5B) & 0.25 & 0.30 & 1.07 & 27.8\% \\
& Qwen-2.5 (0.5B $\to$ 3B) & 0.25 & 0.37 & 1.13 & 32.7\% \\
& Qwen-2.5 (1.5B $\to$ 3B) & 0.28 & 0.33 & 1.13 & 29.1\% \\
\bottomrule
\end{tabular}
}

\caption{Scaling summary for \textsc{MCQA} across source--target model pairs and attribution methods. \textbf{Gold} denotes the direct target-model attribution result, and \textbf{DFA} denotes the strongest aligned result across in-distribution, near-distribution, and zero-shot transfer. Recovery ratio is computed as $\frac{\text{Aligned(best)}}{\text{Gold}} \times 100$, measuring how much target-model faithfulness is recovered by DFA. Values above $100\%$ indicate that the aligned circuit is more faithful than the direct target-model circuit. }
\label{tab:scaling-summary-mcqa}
\end{table*}

\begin{figure}[t]
  \centering
  \includegraphics[width=\linewidth]{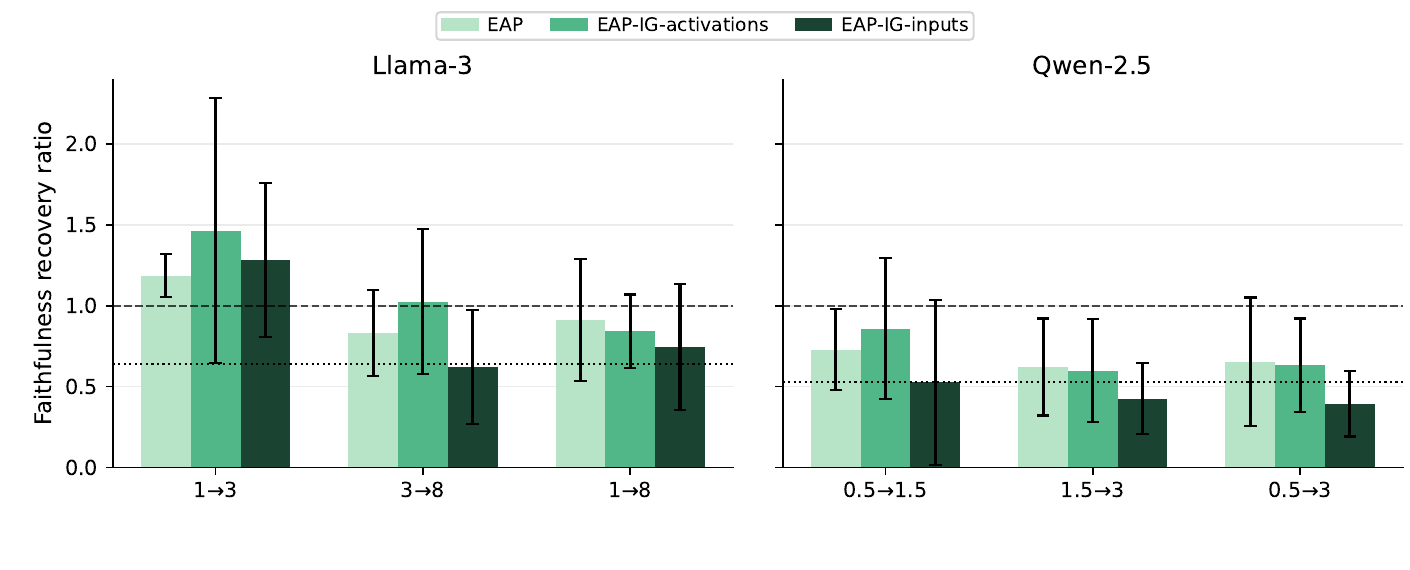}
    \caption{Mean faithfulness recovery ratio across tasks for each attribution method and source--target pair, shown separately for the Llama-3 and Qwen-2.5 families. Recovery ratio is $\text{Aligned(zero-shot)} / \text{Gold}$, where \textbf{Gold} is direct target-model attribution. Bars show the mean across tasks, and error bars denote one standard deviation. The dashed line marks parity with direct target attribution.}
  \label{fig:cross-model-recovery-mean-over-tasks}
\end{figure}

We next test whether DFA remains effective as the source--target gap increases and across model families. Table~\ref{tab:scaling-summary-mcqa} then reports representative quantitative results for \textsc{MCQA}, while Figure~\ref{fig:cross-model-recovery-mean-over-tasks} summarizes zero-shot recovery averaged over tasks.\footnote{Additional results are reported in Appendix Tables~\ref{tab:scaling-summary-mcqa-eap-eap-ig-activations}--\ref{tab:scaling-summary-arithmetic-subtraction}. Best-alignment and near-distribution summaries are also provided in Appendix Figures~\ref{fig:cross-model-recovery-mean-over-tasks-best} and \ref{fig:cross-model-recovery-mean-over-tasks-in-distribution}.}

Two patterns emerge. First, recovery is strongest within the Llama-3 family, especially for the smaller scale gap, where aligned circuits often recover a large fraction of the direct target-model reference and can sometimes be competitive with it. Second, recovery is weaker for Qwen-2.5, particularly for larger gaps, indicating that alignment becomes harder as architectural mismatch grows. Even so, Figure~\ref{fig:cross-model-recovery-mean-over-tasks} shows that for Qwen-2.5, \textsc{NAP} and \textsc{NAP-IG-Activations} remain consistently above the random baseline, indicating non-trivial transfer beyond unlearned alignment. Overall, these results suggest that cross-model circuit alignment is feasible, but becomes more difficult as scale differences increase. As a sanity check, we also evaluate the reverse transfer direction and find it is often easier to map from a larger model to a smaller one than vice versa (Table~\ref{tab:leaderboard-llama3-3b-to-llama3-1b}), consistent with the intuition that compression is easier than extrapolation. Together with weaker Qwen-2.5 results, this suggests that transfer becomes harder with larger scale gaps or less compatible model families; see Appendix~\ref{app:transfer-limits} for further discussion.

\subsection{Cross-task transfer structure}
\label{subsec:exp_heatmap}

\begin{figure}[!t]
  \centering
  \includegraphics[width=\linewidth]{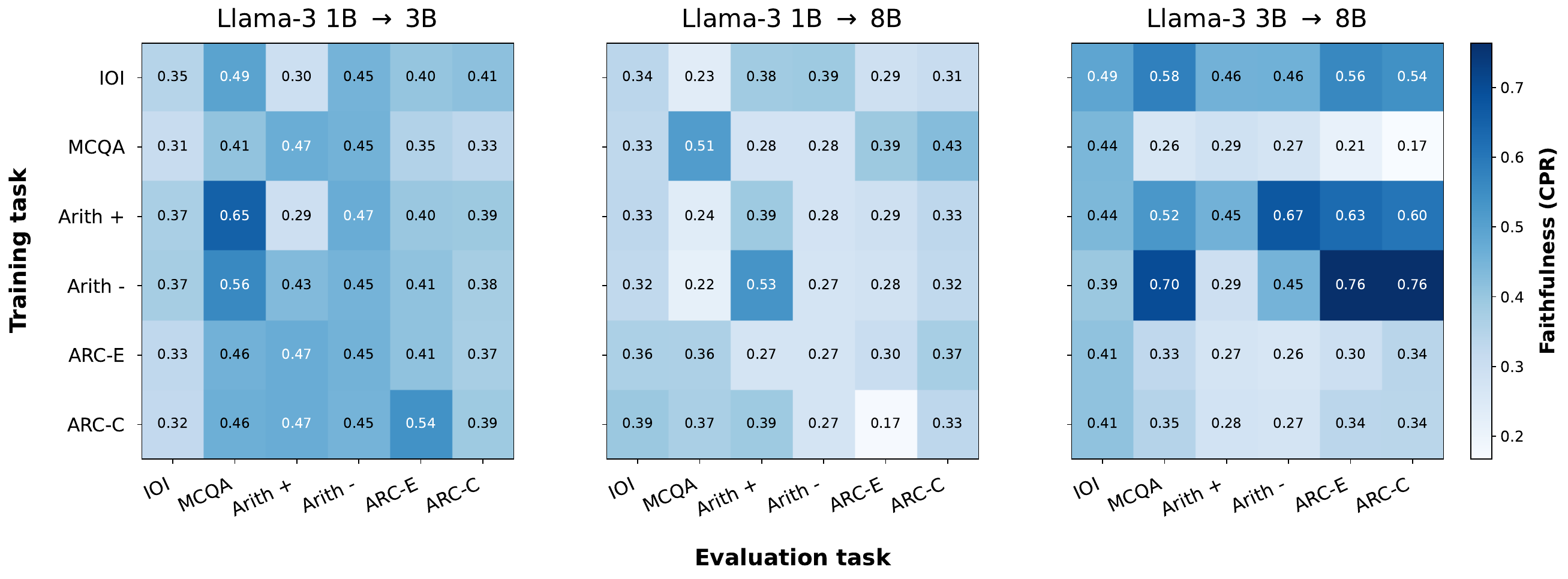}
    \caption{Cross-task transfer matrix for \textsc{NAP-IG-Inputs} on Llama-3, evaluated with CPR. Rows denote training tasks and columns evaluation tasks. Entries report faithfulness of transferred circuits; strong off-diagonal values indicate cross-task generalization.}
  \label{fig:cross-task-heatmap-llama-eap-ig-inputs-llama}
\end{figure}

\begin{figure}[t]
  \centering
  \includegraphics[width=0.9\linewidth]{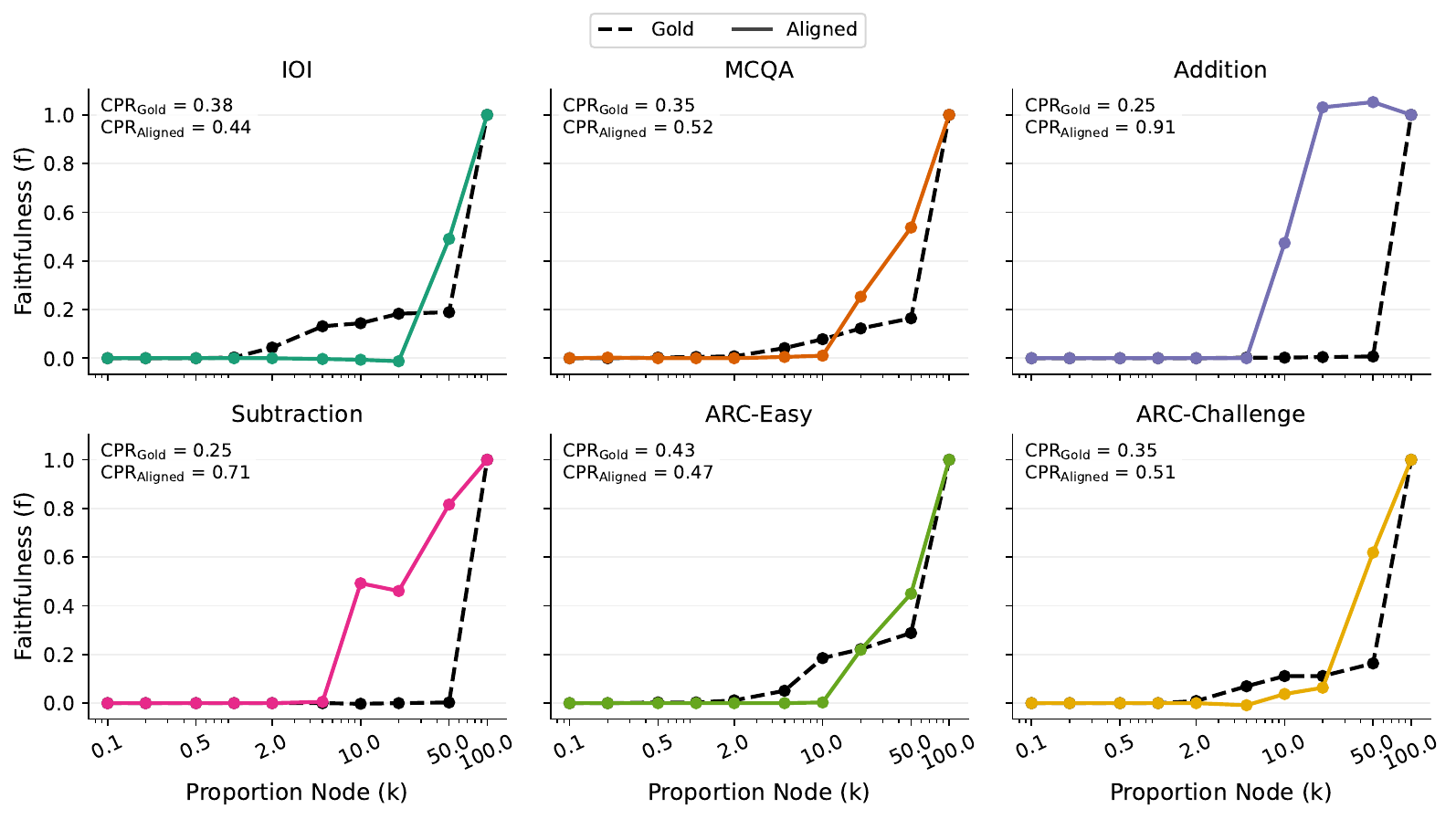}
    \caption{Best aligned circuits (\textbf{Aligned}) versus direct target-model circuits (\textbf{Gold}) across tasks for \textsc{NAP-IG-Activations} on Llama-3 ($1$B$\to3$B). Across tasks, aligned circuits recover similar faithfulness curves and often achieve comparable or higher CPR.}
  \label{fig:1c-eap-ig-act-llama3-1b-3b}
\end{figure}

The main results averages establish that transfer works, but they do not show \emph{which} tasks transfer well to which others. To expose this structure, we train $W$ on one task at a time and evaluate it on every task, yielding the task-to-task transfer matrices in Figure~\ref{fig:cross-task-heatmap-llama-eap-ig-inputs-llama}; additional heatmaps for other methods and model pairs are reported in Appendix Figures~\ref{fig:cross-task-heatmap-llama-eap-ig-activations-llama}--\ref{fig:cross-task-heatmap-llama-eap-qwen}. The matrices show that transfer is structured rather than random: many off-diagonal entries remain strong, so some learned correspondences clearly generalize beyond the training task. At the same time, the diagonal is not always dominant. In particular, tasks such as \textsc{MCQA}, \textsc{ARC-E}, and \textsc{ARC-C} often transfer well across multiple training tasks, while some matched-task entries are only moderate. This suggests that certain training tasks induce more reusable source--target correspondences than others.

Figure~\ref{fig:1c-eap-ig-act-llama3-1b-3b} provides a complementary per-task view for the best aligned setting. Across tasks, the aligned circuits usually recover faithfulness curves similar to the direct target-model circuits and often achieve comparable or higher CPR, especially on \textsc{Addition}, \textsc{Subtraction}, and \textsc{ARC-Challenge}. Taken together, these results suggest that DFA is not merely memorizing isolated task-specific masks. Rather, it learns partially reusable functional correspondences whose transfer strength depends on both task relatedness and the quality of the underlying node-level circuit signal.

\section{Discussion}

We studied whether mechanistic circuits discovered in smaller language models can be transferred to larger models through learned alignment. We introduced \textbf{Differentiable Faithfulness Alignment (DFA)}, which projects source-model circuit scores into the target model and learns this mapping with a differentiable faithfulness objective.

Our results show that smaller-model circuits can often recover faithful target-model circuits without direct attribution on the target model, with strongest performance on Llama-3 $1$B$\rightarrow3$B. Transfer weakens for larger source--target gaps and is less favorable on Qwen-2.5, indicating that cross-model circuit transfer is real but not uniform. Ablations further show that these gains depend on structured source-model signal and a learned source--target correspondence, rather than random masks or architectural priors. In some settings, DFA even yields target-model circuits more faithful than direct node attribution on the target model, suggesting that smaller models provide cleaner mechanistic signals for alignment.

More broadly, our findings suggest that mechanistic generalization may be better studied through explicit functional alignment than through overlap alone.

Our study is limited to node-level circuits, where localization is generally weaker than edge-level methods \citep{mib-2025}. Results are also not uniform across model pairs, with weaker transfer for larger source--target gaps and on Qwen-2.5. Extending DFA to richer circuit representations and broader model families is an important direction for future work.

\section*{Ethics Statement}
This work studies methods for transferring circuit information across language models to improve mechanistic analysis. We do not collect human-subject data or introduce a deployed system. The main ethical considerations are the computational cost of large-model analysis and the possibility that improved interpretability tools could be used in both beneficial and potentially dual-use settings. We therefore aim to support transparency and reproducibility by clearly describing our experimental setup and limitations.

\section*{Acknowledgements}
The authors acknowledge the use of resources provided by the Isambard-AI National AI Research Resource (AIRR). Isambard-AI is operated by the University of Bristol and is funded by the UK Government’s Department for Science, Innovation and Technology (DSIT) via UK Research and Innovation; and the Science and Technology Facilities Council [ST/AIRR/I-A-I/1023]~\citep{mcintoshsmith2024isambardaileadershipclasssupercomputer}. This research was supported by Coefficient Giving, the Israel Science Foundation (grant No. 2942/25), and the European Union (ERC, Control-LM, 101165402). Views and opinions expressed are however those of the author(s) only and do not necessarily reflect those of the European Union or the European Research Council Executive Agency. Neither the European Union nor the granting authority can be held responsible for them.

\bibliography{colm2026_conference}
\bibliographystyle{colm2026_conference}

\clearpage

\appendix

\section{Additional Experimental Statistics}
\label{app:transfer-sizes}

Table~\ref{tab:dataset-splits} reports dataset split sizes. Table~\ref{tab:transfer-training-sizes} reports the training size for each transfer regime. These regimes are not matched-budget comparisons: in-distribution tests same-task alignment, near-distribution tests transfer between related tasks, and zero-shot/leave-one-out tests transfer to a held-out task.

\begin{table}[!htbp]
\centering
\footnotesize
\setlength{\tabcolsep}{8pt}
\begin{tabular}{@{}lrrr@{}}
\toprule
\textbf{Dataset} & \textbf{Train} & \textbf{Validation} & \textbf{Public Test} \\
\midrule
IOI & 10000 & 10000 & 1000 \\
MCQA & 110 & 50 & 50 \\
Arithmetic Addition & 34400 & 4920 & 1000 \\
Arithmetic Subtraction & 17400 & 2484 & 1000 \\
ARC-Easy & 2251 & 570 & 1188 \\
ARC-Challenge & 1119 & 299 & 586 \\
\bottomrule
\end{tabular}
\caption{Dataset split sizes used in our experiments.}
\label{tab:dataset-splits}
\end{table}

\begin{table}[!htbp]
\centering
\footnotesize
\setlength{\tabcolsep}{6pt}
\begin{tabular}{@{}lrrr@{}}
\toprule
\textbf{Evaluation Task} & \textbf{In-Dist.} & \textbf{Near-Dist.} & \textbf{Zero-Shot} \\
\midrule
IOI & 10000 & - & 55280 \\
MCQA & 110 & - & 65170 \\
Arithmetic Addition & 34400 & 17400 & 30880 \\
Arithmetic Subtraction & 17400 & 34400 & 47880 \\
ARC-Easy & 2251 & 1119 & 63029 \\
ARC-Challenge & 1119 & 2251 & 64161 \\
\bottomrule
\end{tabular}
\caption{Training sizes for each transfer setting. Zero-shot trains on all tasks except the evaluation task.}
\label{tab:transfer-training-sizes}
\end{table}

\section{Diagnostic Ablations: Regression and Nonlinear Alignment}
\label{app:diagnostic-ablations}

We run two diagnostic ablations on the primary Llama-3 (1B $\to$ 3B) setting with \textsc{NAP-IG-Inputs}. These experiments are not exhaustive sweeps over all model pairs and attribution methods. Instead, they isolate whether directly regressing to target attribution scores is sufficient, and whether a higher-capacity alignment map improves transfer.

First, we compare against a supervised regression baseline. This baseline learns $\hat{\mathbf{s}}_L = \mathbf{s}_S W$ and optimizes
\[
\mathcal{L}_{\mathrm{reg}}
=
\mathrm{MSE}(\hat{\mathbf{s}}_L,\mathbf{s}_L)
+
\lambda \left\|\sigma(\hat{\mathbf{s}}_L/\tau)\right\|_1 .
\]
Unlike DFA, this baseline uses target-model attribution scores as labels during training and is not trained through target-model causal interventions.

Second, we replace the linear score projection with a two-layer MLP:
\[
\mathbf{h}=\mathrm{GELU}(\mathbf{s}_S W_1+\mathbf{b}_1), \qquad
\mathbf{z}_L=\mathbf{h}W_2+\mathbf{b}_2, \qquad
\mathbf{m}_L=\sigma(\mathbf{z}_L/\tau).
\]
The nonlinear model is trained with the same differentiable faithfulness objective as DFA.

\begin{table*}[!t]
\centering
\resizebox{\textwidth}{!}{%
\begin{tabular}{llccccccc}
\toprule
\textbf{Method Category} & \textbf{Transfer Setting} & \textbf{IOI} & \textbf{MCQA} & \textbf{Arith +} & \textbf{Arith -} & \textbf{ARC-E} & \textbf{ARC-C} & \textbf{Avg.} \\
\midrule
Baseline & Random Alignment & 0.25 & 0.25 & 0.25 & 0.25 & 0.25 & 0.25 & 0.25 \\
\midrule
\multirow{4}{*}{DFA} 
& Gold (Upper Bound) & 0.28 & 0.47 & 0.26 & 0.25 & 0.52 & 0.48 & 0.38 \\
& DFA (In-Distribution) & 0.35 & 0.41 & 0.29 & 0.45 & 0.41 & 0.39 & 0.38 \\
& DFA (Near-Distribution) & - & - & 0.43 & 0.47 & 0.54 & 0.37 & 0.45 \\
& DFA (Far/Zero-Shot) & 0.31 & 0.48 & 0.51 & 0.49 & 0.47 & 0.38 & 0.44 \\
\midrule
\multirow{3}{*}{Nonlinear DFA} 
& In-Distribution & 0.34 & 0.50 & 0.53 & 0.53 & 0.57 & 0.52 & 0.50 \\
& Near-Distribution & - & - & 0.54 & 0.55 & 0.51 & 0.57 & 0.54 \\
& Far/Zero-Shot & 0.33 & 0.27 & 0.40 & 0.40 & 0.27 & 0.27 & 0.32 \\
\midrule
\multirow{3}{*}{Supervised Regression} 
& In-Distribution & 0.25 & 0.25 & 0.26 & 0.25 & 0.43 & 0.40 & 0.31 \\
& Near-Distribution & - & - & 0.25 & 0.25 & 0.42 & 0.41 & 0.33 \\
& Far/Zero-Shot & 0.25 & 0.43 & 0.25 & 0.25 & 0.25 & 0.41 & 0.31 \\
\bottomrule
\end{tabular}%
}
\caption{Diagnostic ablations on Llama-3 (1B $\to$ 3B) with \textsc{NAP-IG-Inputs}. These results isolate the effect of training objective and alignment parameterization; they are not exhaustive sweeps over all model pairs or attribution methods.}
\label{tab:diagnostic-regression-nonlinear}
\end{table*}

The supervised regression baseline underperforms DFA in all three transfer regimes, with average CPR of 0.31/0.33/0.31 compared with DFA's 0.38/0.45/0.44. This suggests that directly matching target attribution scores is less effective than learning the alignment through target-model intervention behavior, even though regression uses target attribution labels that DFA does not require.

The nonlinear model improves in-distribution and near-distribution transfer, raising average CPR from 0.38 to 0.50 and from 0.45 to 0.54, respectively. However, it performs worse in far/zero-shot transfer, dropping from 0.44 to 0.32. This suggests that additional expressivity can help when training and evaluation tasks are closely related, but may reduce task-stable generalization.

\section{Computational Cost and Amortization}
\label{app:compute}

DFA does not make target-model computation free. Training the alignment matrix $W$ still requires clean/corrupted target-model forward passes under the differentiable faithfulness objective. The computational benefit is therefore amortized rather than one-shot. Once $W$ is learned for a source--target model pair, new target-side circuit scores can be predicted from source-model scores using a lightweight projection, without rerunning direct target-model attribution for every new task. This makes DFA most useful when the same source--target alignment can be reused across multiple tasks or behaviors.

\section{Additional Analysis of Transfer Limits}
\label{app:transfer-limits}

DFA is not uniformly reliable across all model pairs. Transfer is strongest for Llama-3 (1B $\to$ 3B) and weaker for larger source--target gaps and for Qwen-2.5. One possible explanation is that larger target models contain richer or more distributed computations, making it harder for a smaller source-model score vector to predict the relevant target-model components. This is consistent with our reverse-transfer sanity check, where mapping from a larger model to a smaller model is often easier than mapping from a smaller model to a larger one.

The weaker Qwen-2.5 results may also reflect differences in node-level organization or noisier node-level attribution signals for that model family. Since DFA aligns node-importance scores, its performance depends on the quality and compatibility of the source and target node-level circuit signals. These results therefore support non-trivial cross-model transfer, but also show that node-level DFA becomes harder as the scale gap or architectural mismatch increases.

\section{Main Results: Zero-Shot, In-Distribution, and Near-Distribution Transfer}

We report the full transfer results for additional source–target model pairs and attribution methods omitted from the main text. Tables \ref{tab:leaderboard-llama3-1b-to-llama3-3b：eap-and-eap-ig-activations}--\ref{tab:leaderboard-qwen2-5-1-5b-to-qwen2-5-3b} extend the main Llama-3 (1B→3B) analysis to larger Llama-3 and Qwen-2.5 pairs, and show performance for NAP, NAP-IG-Activations, and NAP-IG-Inputs under in-distribution, near-distribution, and zero-shot transfer. These results support the same overall trend as the main text: transfer is strongest for smaller within-family gaps and weakens for larger or cross-family gaps.

\begin{table*}[!t]
\centering
\resizebox{\textwidth}{!}{
\begin{tabular}{llccccccc}
\toprule
\textbf{Method Category} & \textbf{Transfer Setting} & \textbf{IOI} & \textbf{MCQA} & \textbf{Arith +} & \textbf{Arith -} & \textbf{ARC-E} & \textbf{ARC-C} & \textbf{Avg.} \\
\midrule
Baseline & Random Alignment & 0.25 & 0.25 & 0.25 & 0.25 & 0.25 & 0.25 & 0.25 \\
\midrule
\multirow{4}{*}{NAP} & Gold (Upper Bound) & 0.25 & 0.35 & 0.25 & 0.25 & 0.37 & 0.34 & 0.30 \\
& DFA (In-Distribution) & \textbf{0.32} & \textbf{0.49} & 0.26 & \underline{0.26} & \textbf{0.46} & \textbf{0.52} & \underline{0.39} \\
& DFA (Near-Distribution) & - & - & \textbf{0.39} & 0.25 & \underline{0.44} & 0.48 & \textbf{0.39} \\
& DFA (Far/Zero-Shot) & \underline{0.27} & \underline{0.46} & \underline{0.28} & \textbf{0.26} & 0.44 & \underline{0.48} & 0.37 \\
\midrule
\multirow{4}{*}{NAP-IG-Activations} & Gold (Upper Bound) & \textbf{0.38} & 0.35 & 0.25 & 0.25 & \textbf{0.43} & 0.35 & 0.33 \\
& DFA (In-Distribution) & 0.33 & \underline{0.39} & 0.53 & \textbf{0.45} & \underline{0.41} & \underline{0.44} & 0.42 \\
& DFA (Near-Distribution) & - & - & \underline{0.72} & 0.28 & 0.41 & 0.37 & \underline{0.44} \\
& DFA (Far/Zero-Shot) & \underline{0.33} & \textbf{0.46} & \textbf{0.82} & \underline{0.28} & 0.34 & \textbf{0.51} & \textbf{0.46} \\
\bottomrule
\end{tabular}
}
\caption{Circuit Recovery Performance (Llama-3 (1B $\to$ 3B)). We report the faithfulness score using CPR. Near-distribution evaluation is only applicable to Arithmetic and ARC task pairs.}
\label{tab:leaderboard-llama3-1b-to-llama3-3b：eap-and-eap-ig-activations}
\end{table*}

\begin{table*}[!t]
\centering
\resizebox{\textwidth}{!}{
\begin{tabular}{llccccccc}
\toprule
\textbf{Method Category} & \textbf{Transfer Setting} & \textbf{IOI} & \textbf{MCQA} & \textbf{Arith +} & \textbf{Arith -} & \textbf{ARC-E} & \textbf{ARC-C} & \textbf{Avg.} \\
\midrule
Baseline & Random Alignment & 0.25 & 0.25 & 0.25 & 0.25 & 0.25 & 0.25 & 0.25 \\
\midrule
\multirow{4}{*}{GPT-2 small $\rightarrow$ GPT-2 medium} 
& Gold (Upper Bound) & \textbf{1.42} & \underline{0.25} & \textbf{0.25} & 0.25 & 0.25 & 0.25 & \textbf{0.45} \\
& DFA (In-Distribution) & 0.22 & \textbf{0.39} & -1.09 & \textbf{0.44} & \textbf{0.33} & \underline{0.31} & 0.10 \\
& DFA (Near-Distribution) & - & - & -0.76 & \underline{0.41} & \underline{0.33} & \textbf{0.34} & 0.08 \\
& DFA (Far/Zero-Shot) & \underline{0.25} & 0.25 & \underline{-0.09} & 0.33 & 0.25 & 0.25 & \underline{0.21} \\
\midrule
\multirow{4}{*}{GPT-2 small $\rightarrow$ GPT-2 large} 
& Gold (Upper Bound) & \textbf{1.30} & 0.25 & 0.25 & 0.25 & 0.30 & \textbf{0.46} & \textbf{0.47} \\
& DFA (In-Distribution) & \underline{0.30} & \textbf{0.35} & 0.50 & 0.35 & \underline{0.30} & \underline{0.30} & 0.35 \\
& DFA (Near-Distribution) & - & - & \underline{0.50} & \underline{0.35} & \textbf{0.35} & 0.29 & \underline{0.37} \\
& DFA (Far/Zero-Shot) & 0.25 & \underline{0.25} & \textbf{0.60} & \textbf{0.54} & 0.25 & 0.25 & 0.36 \\
\midrule
\multirow{4}{*}{GPT-2 medium $\rightarrow$ GPT-2 large} 
& Gold (Upper Bound) & \textbf{1.30} & 0.25 & \underline{0.25} & 0.25 & 0.30 & \underline{0.46} & \textbf{0.47} \\
& DFA (In-Distribution) & \underline{0.27} & \underline{0.37} & \textbf{0.38} & \textbf{0.27} & \underline{0.47} & 0.42 & \underline{0.36} \\
& DFA (Near-Distribution) & - & - & 0.25 & 0.25 & \textbf{0.48} & \textbf{0.48} & 0.36 \\
& DFA (Far/Zero-Shot) & 0.25 & \textbf{0.46} & 0.25 & \underline{0.25} & 0.35 & 0.35 & 0.32 \\
\bottomrule
\end{tabular}
}
\caption{Circuit recovery performance for GPT-2 model families using NAP-IG-Inputs. We report faithfulness using CPR. Near-distribution evaluation is only applicable to Arithmetic and ARC task pairs.}
\label{tab:gpt2_circuit_recovery}
\end{table*}

\begin{table*}[!t]
\centering
\resizebox{\textwidth}{!}{
\begin{tabular}{llccccccc}
\toprule
\textbf{Method Category} & \textbf{Transfer Setting} & \textbf{IOI} & \textbf{MCQA} & \textbf{Arith +} & \textbf{Arith -} & \textbf{ARC-E} & \textbf{ARC-C} & \textbf{Avg.} \\
\midrule
Baseline & Random Alignment & 0.25 & 0.25 & 0.25 & 0.25 & 0.25 & 0.25 & 0.25 \\
\midrule
\multirow{4}{*}{NAP-IG-Inputs} & Gold (Upper Bound) & \textbf{0.63} & \textbf{1.02} & 0.46 & \underline{0.37} & \textbf{0.97} & \textbf{0.94} & \textbf{0.73} \\
& DFA (In-Distribution) & \underline{0.34} & \underline{0.51} & 0.39 & 0.27 & 0.30 & 0.33 & 0.36 \\
& DFA (Near-Distribution) & - & - & \underline{0.53} & 0.28 & 0.17 & 0.37 & 0.34 \\
& DFA (Far/Zero-Shot) & 0.33 & 0.51 & \textbf{0.67} & \textbf{0.39} & \underline{0.47} & \underline{0.38} & \underline{0.46} \\
\midrule
\multirow{4}{*}{NAP} & Gold (Upper Bound) & \underline{0.31} & 0.38 & 0.25 & \underline{0.29} & \textbf{0.57} & \textbf{0.59} & \underline{0.40} \\
& DFA (In-Distribution) & \textbf{0.33} & \textbf{0.46} & \underline{0.35} & \textbf{0.33} & \underline{0.49} & \underline{0.53} & \textbf{0.41} \\
& DFA (Near-Distribution) & - & - & 0.30 & 0.26 & 0.47 & 0.46 & 0.37 \\
& DFA (Far/Zero-Shot) & 0.27 & \underline{0.40} & \textbf{0.40} & 0.26 & 0.30 & 0.29 & 0.32 \\
\midrule
\multirow{4}{*}{NAP-IG-Activations} & Gold (Upper Bound) & 0.35 & \textbf{0.60} & 0.33 & \textbf{0.31} & \textbf{0.59} & \textbf{0.63} & \textbf{0.47} \\
& DFA (In-Distribution) & \textbf{0.43} & 0.32 & 0.38 & 0.27 & 0.29 & 0.27 & 0.33 \\
& DFA (Near-Distribution) & - & - & \textbf{0.41} & 0.27 & 0.25 & 0.27 & 0.30 \\
& DFA (Far/Zero-Shot) & \underline{0.36} & \underline{0.40} & \underline{0.41} & \underline{0.28} & \underline{0.35} & \underline{0.40} & \underline{0.37} \\
\bottomrule
\end{tabular}
}
\caption{Circuit Recovery Performance (Llama-3 (1B $\to$ 8B)). We report the faithfulness score using CPR. Near-distribution evaluation is only applicable to Arithmetic and ARC task pairs.}
\label{tab:leaderboard-llama3-1b-to-llama3-8b}
\end{table*}

\begin{table*}[!t]
\centering
\resizebox{\textwidth}{!}{
\begin{tabular}{llccccccc}
\toprule
\textbf{Method Category} & \textbf{Transfer Setting} & \textbf{IOI} & \textbf{MCQA} & \textbf{Arith +} & \textbf{Arith -} & \textbf{ARC-E} & \textbf{ARC-C} & \textbf{Avg.} \\
\midrule
Baseline & Random Alignment & 0.25 & 0.25 & 0.25 & 0.25 & 0.25 & 0.25 & 0.25 \\
\midrule
\multirow{4}{*}{NAP-IG-Inputs} & Gold (Upper Bound) & \textbf{0.63} & \textbf{1.02} & \textbf{0.46} & 0.37 & \textbf{0.97} & \textbf{0.94} & \textbf{0.73} \\
& DFA (In-Distribution) & 0.49 & 0.26 & \underline{0.45} & 0.45 & 0.30 & 0.34 & 0.38 \\
& DFA (Near-Distribution) & - & - & 0.29 & \textbf{0.67} & \underline{0.34} & \underline{0.34} & \underline{0.41} \\
& DFA (Far/Zero-Shot) & \underline{0.57} & \underline{0.46} & 0.27 & \underline{0.46} & 0.27 & 0.25 & 0.38 \\
\midrule
\multirow{4}{*}{NAP} & Gold (Upper Bound) & \underline{0.31} & \textbf{0.38} & 0.25 & 0.29 & \textbf{0.57} & \textbf{0.59} & \textbf{0.40} \\
& DFA (In-Distribution) & 0.23 & 0.27 & \textbf{0.40} & \textbf{0.40} & 0.29 & 0.28 & 0.31 \\
& DFA (Near-Distribution) & - & - & \underline{0.36} & \underline{0.37} & 0.28 & \underline{0.29} & \underline{0.32} \\
& DFA (Far/Zero-Shot) & \textbf{0.36} & \underline{0.32} & 0.27 & 0.26 & \underline{0.29} & 0.27 & 0.30 \\
\midrule
\multirow{4}{*}{NAP-IG-Activations} & Gold (Upper Bound) & 0.35 & \textbf{0.60} & \underline{0.33} & 0.31 & \textbf{0.59} & \textbf{0.63} & 0.47 \\
& DFA (In-Distribution) & \textbf{0.50} & \underline{0.54} & \textbf{0.46} & 0.53 & 0.53 & 0.56 & \textbf{0.52} \\
& DFA (Near-Distribution) & - & - & 0.27 & \underline{0.59} & \underline{0.56} & \underline{0.56} & \underline{0.49} \\
& DFA (Far/Zero-Shot) & \underline{0.44} & 0.50 & 0.28 & \textbf{0.59} & 0.36 & 0.43 & 0.43 \\
\bottomrule
\end{tabular}
}
\caption{Circuit Recovery Performance (Llama-3 (3B $\to$ 8B)). We report the faithfulness score using CPR. Near-distribution evaluation is only applicable to Arithmetic and ARC task pairs.}
\label{tab:leaderboard-llama3-3b-to-llama3-8b}
\end{table*}

\begin{table*}[!t]
\centering
\resizebox{\textwidth}{!}{
\begin{tabular}{llccccccc}
\toprule
\textbf{Method Category} & \textbf{Transfer Setting} & \textbf{IOI} & \textbf{MCQA} & \textbf{Arith +} & \textbf{Arith -} & \textbf{ARC-E} & \textbf{ARC-C} & \textbf{Avg.} \\
\midrule
Baseline & Random Alignment & 0.25 & 0.25 & - & 0.25 & 0.25 & 0.25 & 0.25 \\
\midrule
\multirow{4}{*}{NAP-IG-Inputs} & Gold (Upper Bound) & \textbf{0.91} & \textbf{1.07} & - & 0.25 & \textbf{0.91} & \textbf{0.95} & \textbf{0.82} \\
& DFA (In-Distribution) & \underline{0.30} & \underline{0.30} & - & \underline{0.27} & 0.29 & \underline{0.29} & \underline{0.29} \\
& DFA (Near-Distribution) & - & - & - & 0.27 & \underline{0.29} & 0.28 & 0.28 \\
& DFA (Far/Zero-Shot) & 0.26 & 0.25 & - & \textbf{0.39} & 0.27 & 0.26 & 0.29 \\
\midrule
\multirow{4}{*}{NAP} & Gold (Upper Bound) & 0.25 & \textbf{0.56} & - & 0.24 & \textbf{0.70} & \textbf{0.67} & \textbf{0.49} \\
& DFA (In-Distribution) & \textbf{0.28} & \underline{0.28} & - & \textbf{0.40} & 0.35 & 0.32 & \underline{0.33} \\
& DFA (Near-Distribution) & - & - & - & 0.25 & 0.30 & 0.35 & 0.30 \\
& DFA (Far/Zero-Shot) & \underline{0.25} & 0.27 & - & \underline{0.25} & \underline{0.38} & \underline{0.36} & 0.31 \\
\midrule
\multirow{4}{*}{NAP-IG-Activations} & Gold (Upper Bound) & 0.25 & \textbf{0.60} & - & 0.24 & \textbf{0.76} & \textbf{0.72} & \textbf{0.51} \\
& DFA (In-Distribution) & \textbf{0.29} & \underline{0.34} & - & \underline{0.40} & 0.32 & 0.31 & 0.33 \\
& DFA (Near-Distribution) & - & - & - & 0.25 & 0.30 & 0.32 & 0.29 \\
& DFA (Far/Zero-Shot) & \underline{0.25} & 0.28 & - & \textbf{0.40} & \underline{0.42} & \underline{0.44} & \underline{0.36} \\
\bottomrule
\end{tabular}
}
\caption{Circuit Recovery Performance (Qwen-2.5 (0.5B $\to$ 1.5B)). We report the faithfulness score using CPR. Near-distribution evaluation is only applicable to Arithmetic and ARC task pairs.}
\label{tab:leaderboard-qwen2-5-0-5b-to-qwen2-5-1-5b}
\end{table*}

\begin{table*}[!t]
\centering
\resizebox{\textwidth}{!}{
\begin{tabular}{llccccccc}
\toprule
\textbf{Method Category} & \textbf{Transfer Setting} & \textbf{IOI} & \textbf{MCQA} & \textbf{Arith +} & \textbf{Arith -} & \textbf{ARC-E} & \textbf{ARC-C} & \textbf{Avg.} \\
\midrule
Baseline & Random Alignment & 0.25 & 0.25 & - & 0.25 & 0.25 & 0.25 & 0.25 \\
\midrule
\multirow{4}{*}{NAP-IG-Inputs} & Gold (Upper Bound) & \textbf{0.91} & \textbf{1.13} & - & \textbf{0.32} & \textbf{1.06} & \textbf{0.94} & \textbf{0.87} \\
& DFA (In-Distribution) & 0.24 & \underline{0.37} & - & \underline{0.26} & 0.34 & \underline{0.34} & 0.31 \\
& DFA (Near-Distribution) & - & - & - & 0.26 & \underline{0.35} & 0.33 & \underline{0.31} \\
& DFA (Far/Zero-Shot) & \underline{0.30} & 0.32 & - & 0.25 & 0.28 & 0.28 & 0.29 \\
\midrule
\multirow{4}{*}{NAP} & Gold (Upper Bound) & 0.25 & \textbf{0.82} & - & \underline{0.25} & \textbf{0.89} & \textbf{1.10} & \textbf{0.66} \\
& DFA (In-Distribution) & \underline{0.25} & \underline{0.37} & - & \textbf{0.26} & \underline{0.37} & 0.38 & \underline{0.33} \\
& DFA (Near-Distribution) & - & - & - & 0.25 & 0.31 & \underline{0.42} & 0.33 \\
& DFA (Far/Zero-Shot) & \textbf{0.32} & 0.27 & - & 0.24 & 0.32 & 0.35 & 0.30 \\
\midrule
\multirow{4}{*}{NAP-IG-Activations} & Gold (Upper Bound) & \underline{0.25} & \textbf{0.90} & - & 0.25 & \textbf{0.90} & \textbf{1.11} & \textbf{0.68} \\
& DFA (In-Distribution) & \textbf{0.25} & \underline{0.36} & - & \textbf{0.26} & 0.29 & 0.30 & 0.29 \\
& DFA (Near-Distribution) & - & - & - & \underline{0.25} & \underline{0.34} & 0.39 & 0.33 \\
& DFA (Far/Zero-Shot) & 0.25 & 0.34 & - & 0.25 & 0.34 & \underline{0.48} & \underline{0.33} \\
\bottomrule
\end{tabular}
}
\caption{Circuit Recovery Performance (Qwen-2.5 (0.5B $\to$ 3B)). We report the faithfulness score using CPR. Near-distribution evaluation is only applicable to Arithmetic and ARC task pairs.}
\label{tab:leaderboard-qwen2-5-0-5b-to-qwen2-5-3b}
\end{table*}

\begin{table*}[!t]
\centering
\resizebox{\textwidth}{!}{
\begin{tabular}{llccccccc}
\toprule
\textbf{Method Category} & \textbf{Transfer Setting} & \textbf{IOI} & \textbf{MCQA} & \textbf{Arith +} & \textbf{Arith -} & \textbf{ARC-E} & \textbf{ARC-C} & \textbf{Avg.} \\
\midrule
Baseline & Random Alignment & 0.25 & 0.28 & - & 0.25 & 0.27 & 0.27 & 0.26 \\
\midrule
\multirow{4}{*}{NAP-IG-Inputs} & Gold (Upper Bound) & \textbf{0.91} & \textbf{1.13} & - & \textbf{0.32} & \textbf{1.06} & \textbf{0.94} & \textbf{0.87} \\
& DFA (In-Distribution) & 0.29 & 0.32 & - & 0.26 & 0.29 & \underline{0.31} & 0.29 \\
& DFA (Near-Distribution) & - & - & - & 0.26 & \underline{0.31} & 0.29 & 0.29 \\
& DFA (Far/Zero-Shot) & \underline{0.36} & \underline{0.33} & - & \underline{0.27} & 0.29 & 0.29 & \underline{0.31} \\
\midrule
\multirow{4}{*}{NAP} & Gold (Upper Bound) & \underline{0.25} & \textbf{0.82} & - & 0.25 & \textbf{0.89} & \textbf{1.10} & \textbf{0.66} \\
& DFA (In-Distribution) & \textbf{0.28} & 0.26 & - & \textbf{0.27} & 0.27 & 0.29 & 0.27 \\
& DFA (Near-Distribution) & - & - & - & \underline{0.26} & 0.29 & 0.27 & 0.27 \\
& DFA (Far/Zero-Shot) & 0.23 & \underline{0.28} & - & 0.26 & \underline{0.40} & \underline{0.38} & \underline{0.31} \\
\midrule
\multirow{4}{*}{NAP-IG-Activations} & Gold (Upper Bound) & \underline{0.25} & \textbf{0.90} & - & 0.25 & \textbf{0.90} & \textbf{1.11} & \textbf{0.68} \\
& DFA (In-Distribution) & \textbf{0.29} & 0.25 & - & \textbf{0.27} & 0.27 & 0.29 & 0.28 \\
& DFA (Near-Distribution) & - & - & - & 0.26 & 0.29 & 0.27 & 0.27 \\
& DFA (Far/Zero-Shot) & 0.23 & \underline{0.30} & - & \underline{0.26} & \underline{0.35} & \underline{0.34} & \underline{0.30} \\
\bottomrule
\end{tabular}
}
\caption{Circuit Recovery Performance (Qwen-2.5 (1.5B $\to$ 3B)). We report the faithfulness score using CPR. Near-distribution evaluation is only applicable to Arithmetic and ARC task pairs.}
\label{tab:leaderboard-qwen2-5-1-5b-to-qwen2-5-3b}
\end{table*}

\FloatBarrier

\subsection{Sanity Check}

As a sanity check, we also evaluate reverse transfer from larger models to smaller ones. Table \ref{tab:leaderboard-llama3-3b-to-llama3-1b} reports this setting for Llama-3 3B→1B and tests whether the asymmetry discussed in the main text is borne out quantitatively. Consistent with the main text, this reverse direction is often easier than small-to-large transfer.

\begin{table*}[!t]
\centering

\resizebox{\textwidth}{!}{
\begin{tabular}{llccccccc}

\toprule
\textbf{Method} & \textbf{Transfer Setting} & \textbf{IOI} & \textbf{MCQA} & \textbf{Arith +} & \textbf{Arith -} & \textbf{ARC-E} & \textbf{ARC-C} & \textbf{Avg.} \\
\midrule
Baseline & Random Alignment & 0.25 & 0.25 & 0.25 & 0.25 & 0.25 & 0.25 & 0.25 \\
\midrule
\multirow{4}{*}{EAP-IG-Inputs} & Gold (Upper Bound) & 0.33 & \underline{0.42} & 0.25 & 0.25 & 0.32 & 0.31 & 0.31 \\
& Ours (In-Distribution) & \textbf{0.61} & \textbf{0.49} & 0.36 & 0.38 & \textbf{0.49} & \underline{0.46} & \underline{0.46} \\
& Ours (Near-Distribution) & - & - & \textbf{0.50} & \textbf{0.47} & \underline{0.46} & \textbf{0.49} & \textbf{0.48} \\
& Ours (Far/Zero-Shot) & \underline{0.49} & 0.39 & \underline{0.44} & \underline{0.45} & 0.43 & 0.45 & 0.44 \\

\bottomrule

\end{tabular}
}

\caption{Validation and ablation study for target model LLaMA-3 1B with source model Llama-3 3B. We report CPR under four structural controls, together with zero-shot and best alignment results. The controls test whether success depends on task difficulty, the topology of the learned mapping, and the semantic structure of the source-model signal.}

\label{tab:leaderboard-llama3-3b-to-llama3-1b}
\end{table*}

\FloatBarrier

\subsection{Additional qualitative visualizations}

\begin{figure}[t]
  \centering
  \includegraphics[width=\linewidth]{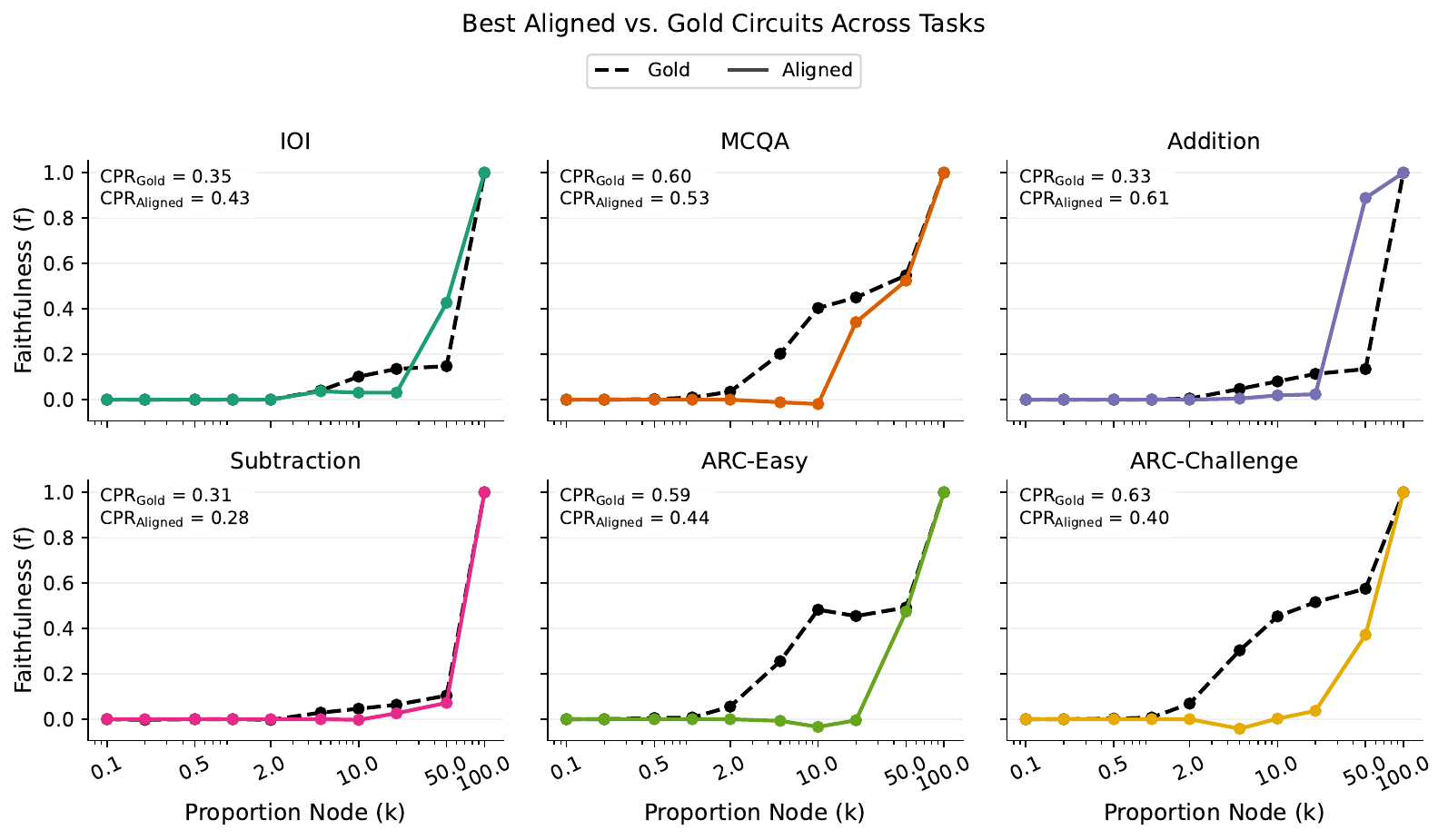}
    \caption{Per-task comparison of direct target-model circuits (\textbf{Gold}) and the DFA-predicted circuits (\textbf{Aligned}) for \textsc{EAP-IG-activations} on (LLaMA-3 (1B $\to$ 8B)). Each panel corresponds to one evaluation task. The Gold curve is obtained by running direct attribution on the target model for that task, while the Aligned curve reports best alignment across in-distribution, near-distribution, and zero-shot transfer. The x-axis shows the proportion of retained nodes $k$, and the y-axis shows faithfulness $f$. CPR denotes the area under the faithfulness curve. Across tasks, the aligned circuits often recover a similar qualitative faithfulness profile to the direct target circuits, and in several cases achieve comparable or stronger CPR.}
  \label{fig:1c-eap-ig-act-llama3-1b-8b}
\end{figure}

\begin{figure}[t]
  \centering
  \includegraphics[width=\linewidth]{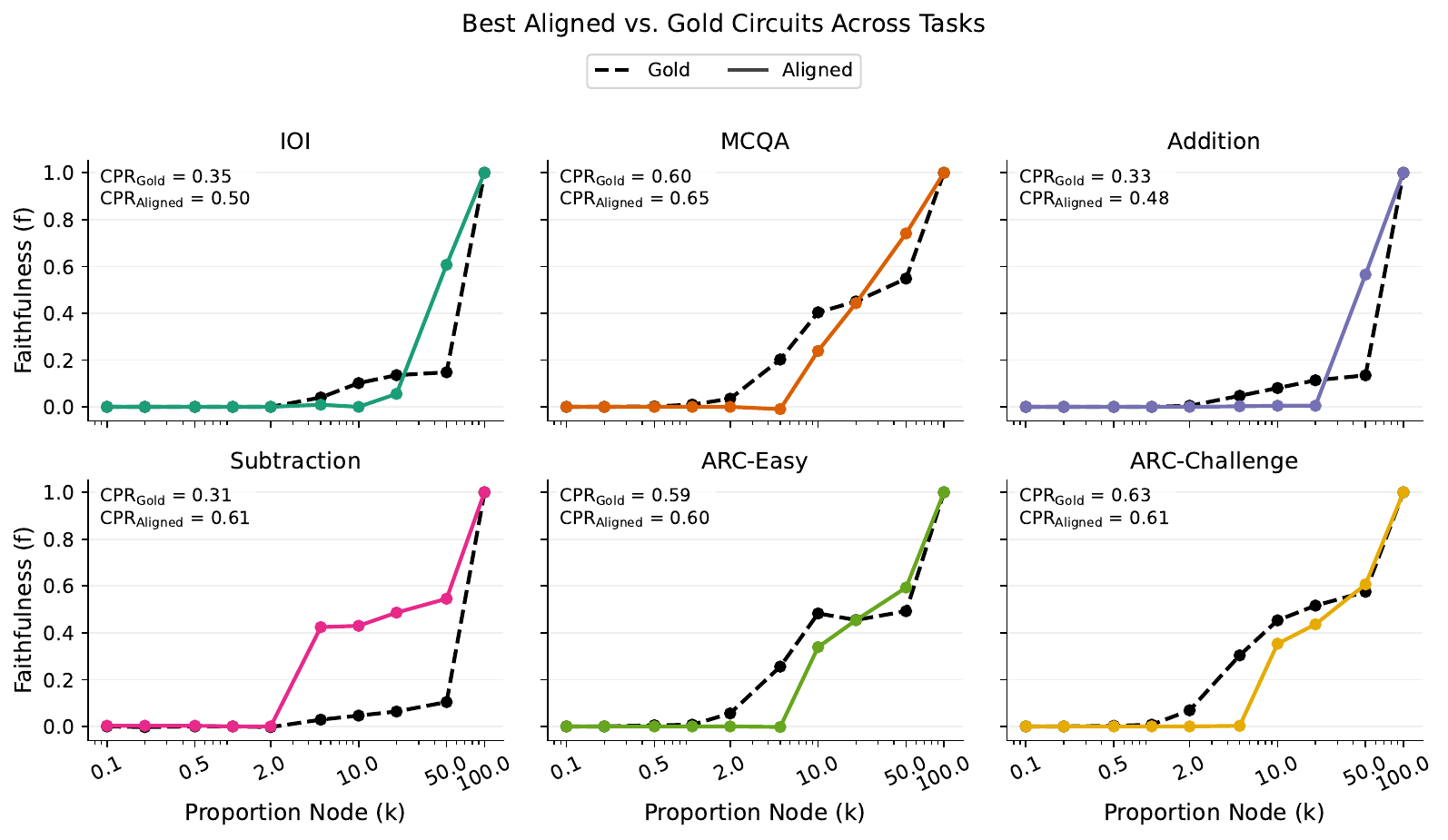}
    \caption{Per-task comparison of direct target-model circuits (\textbf{Gold}) and the DFA-predicted circuits (\textbf{Aligned}) for \textsc{EAP-IG-activations} on (LLaMA-3 (3B $\to$ 8B)). Each panel corresponds to one evaluation task. The Gold curve is obtained by running direct attribution on the target model for that task, while the Aligned curve reports best alignment across in-distribution, near-distribution, and zero-shot transfer. The x-axis shows the proportion of retained nodes $k$, and the y-axis shows faithfulness $f$. CPR denotes the area under the faithfulness curve. Across tasks, the aligned circuits often recover a similar qualitative faithfulness profile to the direct target circuits, and in several cases achieve comparable or stronger CPR.}
  \label{fig:1c-eap-ig-act-llama3-3b-8b}
\end{figure}

\begin{figure}[t]
  \centering
  \includegraphics[width=\linewidth]{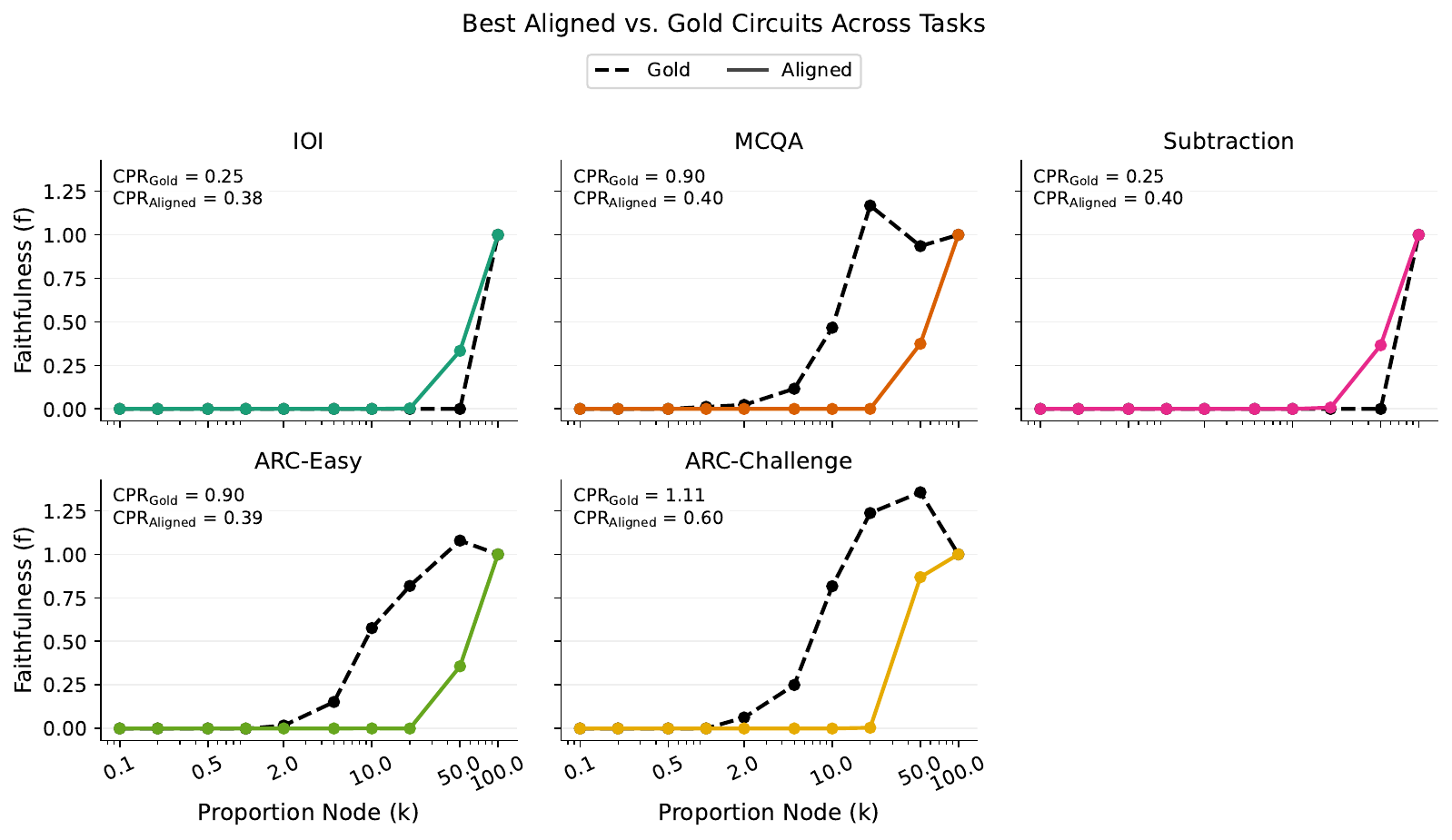}
    \caption{Per-task comparison of direct target-model circuits (\textbf{Gold}) and the DFA-predicted circuits (\textbf{Aligned}) for \textsc{EAP-IG-activations} on (QWEN-2.5 (0.5B $\to$ 3B)). Each panel corresponds to one evaluation task. The Gold curve is obtained by running direct attribution on the target model for that task, while the Aligned curve reports best alignment across in-distribution, near-distribution, and zero-shot transfer. The x-axis shows the proportion of retained nodes $k$, and the y-axis shows faithfulness $f$. CPR denotes the area under the faithfulness curve. Across tasks, the aligned circuits often recover a similar qualitative faithfulness profile to the direct target circuits, and in several cases achieve comparable or stronger CPR.}
  \label{fig:1c-eap-ig-act-qwen2.5-0.5b-3b}
\end{figure}

\begin{figure}[t]
  \centering
  \includegraphics[width=\linewidth]{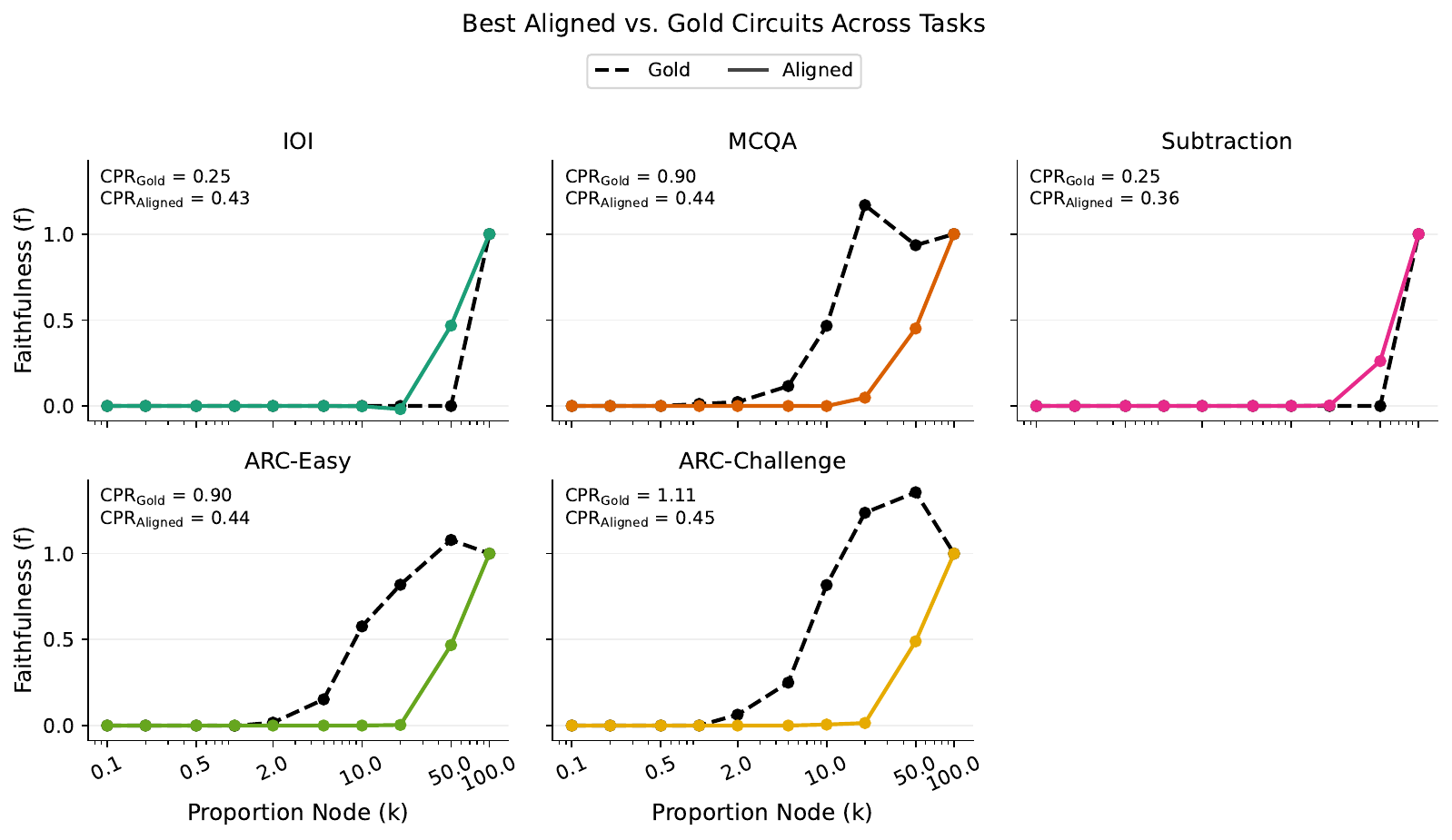}
    \caption{Per-task comparison of direct target-model circuits (\textbf{Gold}) and the DFA-predicted circuits (\textbf{Aligned}) for \textsc{EAP-IG-activations} on (QWEN-2.5 (1.5B $\to$ 3B)). Each panel corresponds to one evaluation task. The Gold curve is obtained by running direct attribution on the target model for that task, while the Aligned curve reports best alignment across in-distribution, near-distribution, and zero-shot transfer. The x-axis shows the proportion of retained nodes $k$, and the y-axis shows faithfulness $f$. CPR denotes the area under the faithfulness curve. Across tasks, the aligned circuits often recover a similar qualitative faithfulness profile to the direct target circuits, and in several cases achieve comparable or stronger CPR.}
  \label{fig:1c-eap-ig-act-qwen2.5-1.5b-3b}
\end{figure}

\begin{figure}[t]
  \centering
  \includegraphics[width=\linewidth]{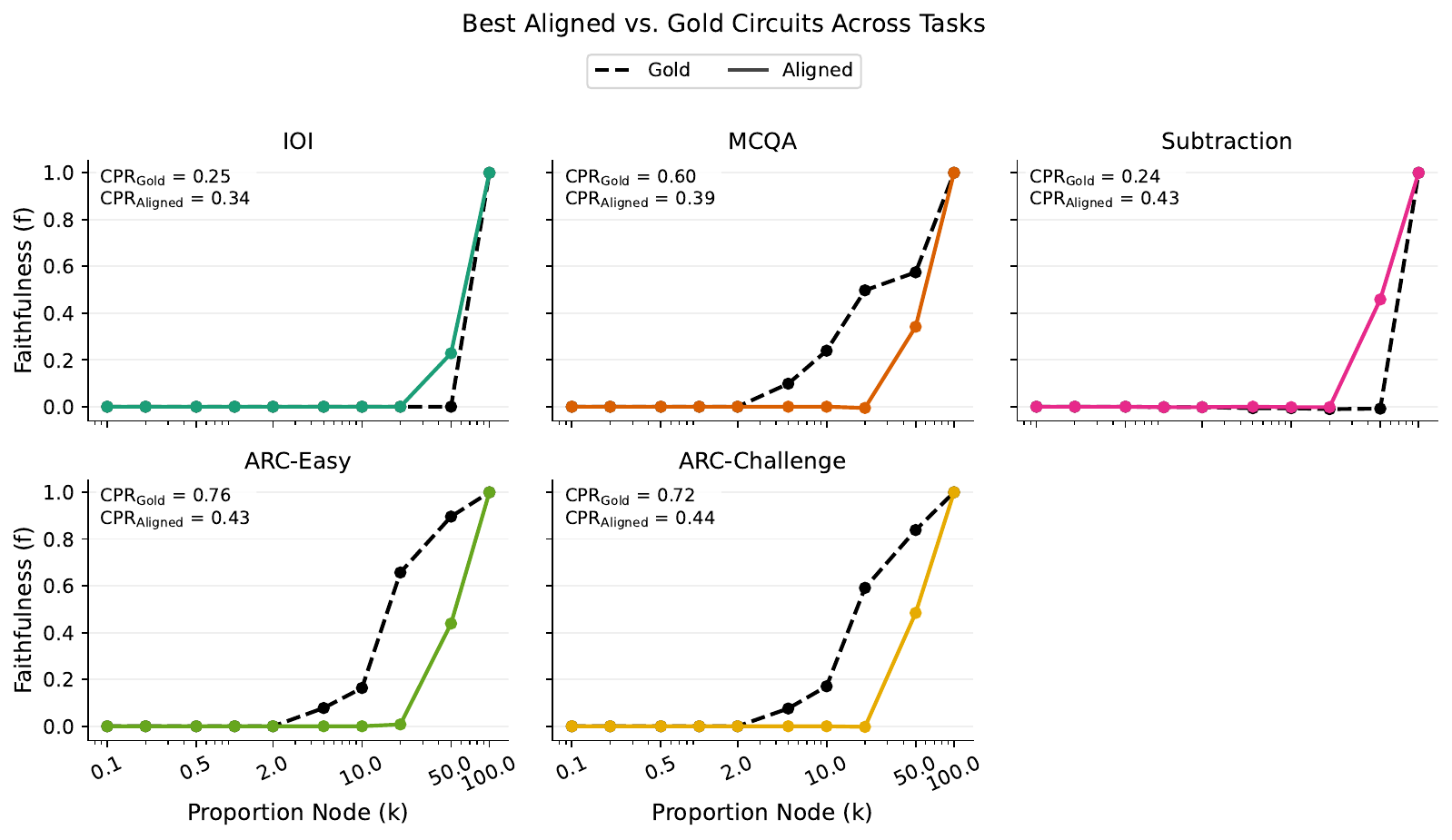}
    \caption{Per-task comparison of direct target-model circuits (\textbf{Gold}) and the DFA-predicted circuits (\textbf{Aligned}) for \textsc{NAP-IG-activations} on (QWEN-2.5 (0.5B $\to$ 1.5B)). Each panel corresponds to one evaluation task. The Gold curve is obtained by running direct attribution on the target model for that task, while the Aligned curve reports best alignment across in-distribution, near-distribution, and zero-shot transfer. The x-axis shows the proportion of retained nodes $k$, and the y-axis shows faithfulness $f$. CPR denotes the area under the faithfulness curve. Across tasks, the aligned circuits often recover a similar qualitative faithfulness profile to the direct target circuits, and in several cases achieve comparable or stronger CPR.}
  \label{fig:1c-eap-ig-act-qwen2.5-0.5b-1.5b}
\end{figure}

Figures \ref{fig:1c-eap-ig-act-llama3-1b-8b}--\ref{fig:1c-eap-ig-act-qwen2.5-0.5b-1.5b} provide per-task comparisons between gold target-model circuits and the best aligned circuits for additional model pairs. These plots complement the aggregate CPR tables by showing that aligned circuits often recover similar faithfulness profiles even when absolute recovery differs across families and scale gaps.

\FloatBarrier

\section{Validating the Alignment Mechanism}

Tables \ref{tab:ablations-summary-llama3-1b-to-llama3-3b-eap-eap-ig-activations}--\ref{tab:ablations-summary-qwen2-5-1-5b-to-qwen2-5-3b} test whether gains depend on learned alignment rather than random masks, scrambled source signals, or simple architectural priors. That is exactly what the main text says Table \ref{tab:ablations-summary-llama3-1b-to-llama3-3b} is testing, extended to more model pairs.

\begin{table*}[!t]
\centering
\resizebox{\textwidth}{!}{
\begin{tabular}{llcccccc}
\toprule
\textbf{Method} & \textbf{Setting / Control} & \textbf{IOI} & \textbf{MCQA} & \textbf{Arith +} & \textbf{Arith -} & \textbf{ARC-E} & \textbf{ARC-C} \\
\midrule
\multirow{6}{*}{NAP} & 1. Random $W$ (Lower Bound) & 0.25 & 0.25 & 0.25 & \underline{0.25} & 0.27 & 0.28 \\
& 2. Scrambled input $s$ & 0.27 & 0.25 & 0.25 & \underline{0.25} & 0.25 & 0.32 \\
& 3. Permuted $W$ columns & \underline{0.29} & 0.27 & 0.25 & \underline{0.25} & 0.41 & 0.31 \\
& 4. Heuristic Depth Mean & 0.25 & 0.25 & 0.25 & \underline{0.25} & 0.25 & 0.25 \\
\addlinespace[4pt]
& \textbf{DFA (Zero-shot)} & 0.27 & \underline{0.46} & \underline{0.28} & \textbf{0.26} & \underline{0.44} & \underline{0.48} \\
& \textbf{DFA (Best)} & \textbf{0.32} & \textbf{0.49} & \textbf{0.39} & \textbf{0.26} & \textbf{0.46} & \textbf{0.52} \\
\midrule
\multirow{6}{*}{NAP-IG-Activations} & 1. Random $W$ (Lower Bound) & 0.25 & 0.25 & 0.25 & 0.25 & 0.26 & 0.25 \\
& 2. Scrambled input $s$ & \underline{0.29} & \underline{0.31} & \underline{0.33} & 0.25 & 0.34 & \underline{0.37} \\
& 3. Permuted $W$ columns & 0.26 & 0.26 & 0.25 & 0.25 & 0.26 & 0.27 \\
& 4. Heuristic Depth Mean & 0.25 & 0.25 & 0.25 & 0.25 & 0.25 & 0.25 \\
\addlinespace[4pt]
& \textbf{DFA (Zero-shot)} & \textbf{0.33} & \textbf{0.46} & \textbf{0.82} & \underline{0.28} & \underline{0.34} & \textbf{0.51} \\
& \textbf{DFA (Best)} & \textbf{0.33} & \textbf{0.46} & \textbf{0.82} & \textbf{0.45} & \textbf{0.41} & \textbf{0.51} \\
\bottomrule
\end{tabular}
}
\caption{Validation and Ablation Study for target model llama3-3b (source llama3-1b) by NAP and NAP-IG-Activations. We report the zero-shot faithfulness drop under various structural corruptions using CPR.}
\label{tab:ablations-summary-llama3-1b-to-llama3-3b-eap-eap-ig-activations}
\end{table*}

\begin{table*}[!t]
\centering
\resizebox{\textwidth}{!}{
\begin{tabular}{llcccccc}
\toprule
\textbf{Method} & \textbf{Setting / Control} & \textbf{IOI} & \textbf{MCQA} & \textbf{Arith +} & \textbf{Arith -} & \textbf{ARC-E} & \textbf{ARC-C} \\
\midrule
\multirow{6}{*}{NAP-IG-Inputs} & 1. Random $W$ (Lower Bound) & 0.25 & 0.25 & 0.25 & \underline{0.25} & 0.25 & 0.25 \\
& 2. Scrambled input $s$ & \underline{0.33} & 0.26 & \underline{0.27} & 0.25 & \underline{0.34} & 0.27 \\
& 3. Permuted $W$ columns & 0.29 & 0.27 & 0.25 & 0.25 & 0.26 & \underline{0.28} \\
& 4. Heuristic Depth Mean & - & - & - & - & - & - \\
\addlinespace[4pt]
& \textbf{DFA (Zero-shot)} & 0.33 & \underline{0.51} & \textbf{0.67} & \textbf{0.39} & \textbf{0.47} & \textbf{0.38} \\
& \textbf{DFA (Best)} & \textbf{0.34} & \textbf{0.51} & \textbf{0.67} & \textbf{0.39} & \textbf{0.47} & \textbf{0.38} \\
\midrule
\multirow{6}{*}{NAP} & 1. Random $W$ (Lower Bound) & 0.25 & 0.25 & 0.25 & 0.25 & 0.25 & 0.25 \\
& 2. Scrambled input $s$ & \textbf{0.39} & 0.25 & \underline{0.26} & 0.25 & 0.25 & 0.30 \\
& 3. Permuted $W$ columns & 0.31 & 0.25 & 0.25 & 0.25 & \underline{0.33} & \underline{0.31} \\
& 4. Heuristic Depth Mean & - & - & - & - & - & - \\
\addlinespace[4pt]
& \textbf{DFA (Zero-shot)} & 0.27 & \underline{0.40} & \textbf{0.40} & \underline{0.26} & 0.30 & 0.29 \\
& \textbf{DFA (Best)} & \underline{0.33} & \textbf{0.46} & \textbf{0.40} & \textbf{0.33} & \textbf{0.49} & \textbf{0.53} \\
\midrule
\multirow{6}{*}{NAP-IG-Activations} & 1. Random $W$ (Lower Bound) & 0.25 & 0.25 & 0.25 & 0.25 & 0.25 & 0.25 \\
& 2. Scrambled input $s$ & \textbf{0.56} & \underline{0.34} & 0.31 & \underline{0.25} & \underline{0.33} & 0.26 \\
& 3. Permuted $W$ columns & 0.30 & 0.26 & 0.25 & \underline{0.25} & 0.26 & \underline{0.27} \\
& 4. Heuristic Depth Mean & - & - & - & - & - & - \\
\addlinespace[4pt]
& \textbf{DFA (Zero-shot)} & 0.36 & \textbf{0.40} & \underline{0.41} & \textbf{0.28} & \textbf{0.35} & \textbf{0.40} \\
& \textbf{DFA (Best)} & \underline{0.43} & \textbf{0.40} & \textbf{0.41} & \textbf{0.28} & \textbf{0.35} & \textbf{0.40} \\
\bottomrule
\end{tabular}
}
\caption{Validation and Ablation Study for target model llama3-8b (source llama3-1b). We report the zero-shot faithfulness drop under various structural corruptions using CPR.}
\label{tab:ablations-summary-llama3-1b-to-llama3-8b}
\end{table*}

\begin{table*}[!t]
\centering
\resizebox{\textwidth}{!}{
\begin{tabular}{llcccccc}
\toprule
\textbf{Method} & \textbf{Setting / Control} & \textbf{IOI} & \textbf{MCQA} & \textbf{Arith +} & \textbf{Arith -} & \textbf{ARC-E} & \textbf{ARC-C} \\
\midrule
\multirow{6}{*}{NAP-IG-Inputs} & 1. Random $W$ (Lower Bound) & 0.25 & 0.25 & 0.25 & 0.25 & 0.25 & 0.25 \\
& 2. Scrambled input $s$ & \underline{0.28} & \underline{0.28} & 0.25 & 0.27 & \underline{0.29} & \textbf{0.49} \\
& 3. Permuted $W$ columns & 0.25 & 0.26 & 0.25 & 0.25 & 0.27 & 0.26 \\
& 4. Heuristic Depth Mean & - & - & - & - & - & - \\
\addlinespace[4pt]
& \textbf{DFA (Zero-shot)} & \textbf{0.57} & \textbf{0.46} & \underline{0.27} & \underline{0.46} & 0.27 & 0.25 \\
& \textbf{DFA (Best)} & \textbf{0.57} & \textbf{0.46} & \textbf{0.45} & \textbf{0.67} & \textbf{0.34} & \underline{0.34} \\
\midrule
\multirow{6}{*}{NAP} & 1. Random $W$ (Lower Bound) & 0.25 & 0.25 & 0.25 & 0.25 & 0.25 & 0.25 \\
& 2. Scrambled input $s$ & \underline{0.26} & 0.26 & 0.25 & 0.26 & 0.25 & \textbf{0.35} \\
& 3. Permuted $W$ columns & 0.25 & \underline{0.26} & 0.25 & 0.25 & \underline{0.26} & 0.26 \\
& 4. Heuristic Depth Mean & - & - & - & - & - & - \\
\addlinespace[4pt]
& \textbf{DFA (Zero-shot)} & \textbf{0.36} & \textbf{0.32} & \underline{0.27} & \underline{0.26} & \textbf{0.29} & 0.27 \\
& \textbf{DFA (Best)} & \textbf{0.36} & \textbf{0.32} & \textbf{0.40} & \textbf{0.40} & \textbf{0.29} & \underline{0.29} \\
\midrule
\multirow{6}{*}{NAP-IG-Activations} & 1. Random $W$ (Lower Bound) & 0.25 & 0.26 & 0.25 & 0.25 & 0.26 & 0.26 \\
& 2. Scrambled input $s$ & 0.25 & 0.31 & 0.25 & \underline{0.42} & 0.25 & 0.29 \\
& 3. Permuted $W$ columns & 0.29 & 0.26 & 0.25 & 0.25 & 0.26 & 0.32 \\
& 4. Heuristic Depth Mean & - & - & - & - & - & - \\
\addlinespace[4pt]
& \textbf{DFA (Zero-shot)} & \underline{0.44} & \underline{0.50} & \underline{0.28} & \textbf{0.59} & \underline{0.36} & \underline{0.43} \\
& \textbf{DFA (Best)} & \textbf{0.50} & \textbf{0.54} & \textbf{0.46} & \textbf{0.59} & \textbf{0.56} & \textbf{0.56} \\
\bottomrule
\end{tabular}
}
\caption{Validation and Ablation Study for target model llama3-8b (source llama3-3b). We report the zero-shot faithfulness drop under various structural corruptions using CPR.}
\label{tab:ablations-summary-llama3-3b-to-llama3-8b}
\end{table*}

\begin{table*}[!t]
\centering
\resizebox{\textwidth}{!}{
\begin{tabular}{llcccccc}
\toprule
\textbf{Method} & \textbf{Setting / Control} & \textbf{IOI} & \textbf{MCQA} & \textbf{Arith +} & \textbf{Arith -} & \textbf{ARC-E} & \textbf{ARC-C} \\
\midrule
\multirow{6}{*}{NAP-IG-Inputs} & 1. Random $W$ (Lower Bound) & 0.25 & 0.25 & - & 0.25 & 0.25 & 0.25 \\
& 2. Scrambled input $s$ & 0.25 & 0.27 & - & 0.25 & 0.25 & 0.27 \\
& 3. Permuted $W$ columns & 0.25 & \textbf{0.32} & - & \underline{0.25} & \underline{0.29} & \underline{0.29} \\
& 4. Heuristic Depth Mean & - & - & - & - & - & - \\
\addlinespace[4pt]
& \textbf{DFA (Zero-shot)} & \underline{0.26} & 0.25 & - & \textbf{0.39} & 0.27 & 0.26 \\
& \textbf{DFA (Best)} & \textbf{0.30} & \underline{0.30} & - & \textbf{0.39} & \textbf{0.29} & \textbf{0.29} \\
\midrule
\multirow{6}{*}{NAP} & 1. Random $W$ (Lower Bound) & 0.25 & 0.25 & - & 0.25 & 0.27 & \underline{0.27} \\
& 2. Scrambled input $s$ & 0.25 & \textbf{0.44} & - & \textbf{0.41} & \underline{0.28} & 0.25 \\
& 3. Permuted $W$ columns & 0.24 & \underline{0.28} & - & 0.25 & 0.27 & 0.26 \\
& 4. Heuristic Depth Mean & - & - & - & - & - & - \\
\addlinespace[4pt]
& \textbf{DFA (Zero-shot)} & \underline{0.25} & 0.27 & - & 0.25 & \textbf{0.38} & \textbf{0.36} \\
& \textbf{DFA (Best)} & \textbf{0.28} & 0.28 & - & \underline{0.40} & \textbf{0.38} & \textbf{0.36} \\
\midrule
\multirow{6}{*}{NAP-IG-Activations} & 1. Random $W$ (Lower Bound) & 0.25 & 0.25 & - & 0.25 & 0.28 & \underline{0.30} \\
& 2. Scrambled input $s$ & 0.25 & \textbf{0.43} & - & 0.25 & \underline{0.33} & 0.25 \\
& 3. Permuted $W$ columns & 0.25 & 0.25 & - & \underline{0.25} & 0.26 & 0.25 \\
& 4. Heuristic Depth Mean & - & - & - & - & - & - \\
\addlinespace[4pt]
& \textbf{DFA (Zero-shot)} & \underline{0.25} & 0.28 & - & \textbf{0.40} & \textbf{0.42} & \textbf{0.44} \\
& \textbf{DFA (Best)} & \textbf{0.29} & \underline{0.34} & - & \textbf{0.40} & \textbf{0.42} & \textbf{0.44} \\
\bottomrule
\end{tabular}
}
\caption{Validation and Ablation Study for target model qwen2.5-1.5b (source qwen2.5-0.5b). We report the zero-shot faithfulness drop under various structural corruptions using CPR.}
\label{tab:ablations-summary-qwen2-5-0-5b-to-qwen2-5-1-5b}
\end{table*}

\begin{table*}[!t]
\centering
\resizebox{\textwidth}{!}{
\begin{tabular}{llcccccc}
\toprule
\textbf{Method} & \textbf{Setting / Control} & \textbf{IOI} & \textbf{MCQA} & \textbf{Arith +} & \textbf{Arith -} & \textbf{ARC-E} & \textbf{ARC-C} \\
\midrule
\multirow{6}{*}{NAP-IG-Inputs} & 1. Random $W$ (Lower Bound) & 0.25 & 0.25 & - & 0.25 & 0.25 & 0.25 \\
& 2. Scrambled input $s$ & 0.25 & 0.36 & - & 0.25 & 0.25 & 0.29 \\
& 3. Permuted $W$ columns & \underline{0.27} & \textbf{0.38} & - & 0.25 & \textbf{0.35} & \textbf{0.34} \\
& 4. Heuristic Depth Mean & - & - & - & - & - & - \\
\addlinespace[4pt]
& \textbf{DFA (Zero-shot)} & \textbf{0.30} & 0.32 & - & \underline{0.25} & 0.28 & 0.28 \\
& \textbf{DFA (Best)} & \textbf{0.30} & \underline{0.37} & - & \textbf{0.26} & \underline{0.35} & \underline{0.34} \\
\midrule
\multirow{6}{*}{NAP} & 1. Random $W$ (Lower Bound) & 0.25 & 0.27 & - & 0.25 & 0.25 & 0.25 \\
& 2. Scrambled input $s$ & 0.25 & \textbf{0.57} & - & \underline{0.25} & 0.29 & 0.25 \\
& 3. Permuted $W$ columns & \underline{0.29} & 0.30 & - & 0.25 & 0.30 & 0.29 \\
& 4. Heuristic Depth Mean & - & - & - & - & - & - \\
\addlinespace[4pt]
& \textbf{DFA (Zero-shot)} & \textbf{0.32} & 0.27 & - & 0.24 & \underline{0.32} & \underline{0.35} \\
& \textbf{DFA (Best)} & \textbf{0.32} & \underline{0.37} & - & \textbf{0.26} & \textbf{0.37} & \textbf{0.42} \\
\midrule
\multirow{6}{*}{NAP-IG-Activations} & 1. Random $W$ (Lower Bound) & 0.25 & 0.27 & - & 0.25 & 0.25 & 0.25 \\
& 2. Scrambled input $s$ & 0.25 & \textbf{0.60} & - & 0.25 & 0.24 & 0.25 \\
& 3. Permuted $W$ columns & \textbf{0.30} & 0.30 & - & \underline{0.25} & 0.31 & \underline{0.31} \\
& 4. Heuristic Depth Mean & - & - & - & - & - & - \\
\addlinespace[4pt]
& \textbf{DFA (Zero-shot)} & 0.25 & 0.34 & - & 0.25 & \underline{0.34} & \textbf{0.48} \\
& \textbf{DFA (Best)} & \underline{0.25} & \underline{0.36} & - & \textbf{0.26} & \textbf{0.34} & \textbf{0.48} \\
\bottomrule
\end{tabular}
}
\caption{Validation and Ablation Study for target model qwen2.5-3b (source qwen2.5-0.5b). We report the zero-shot faithfulness drop under various structural corruptions using CPR.}
\label{tab:ablations-summary-qwen2-5-0-5b-to-qwen2-5-3b}
\end{table*}

\begin{table*}[!t]
\centering
\resizebox{\textwidth}{!}{
\begin{tabular}{llcccccc}
\toprule
\textbf{Method} & \textbf{Setting / Control} & \textbf{IOI} & \textbf{MCQA} & \textbf{Arith +} & \textbf{Arith -} & \textbf{ARC-E} & \textbf{ARC-C} \\
\midrule
\multirow{6}{*}{NAP-IG-Inputs} & 1. Random $W$ (Lower Bound) & 0.25 & 0.28 & - & 0.25 & 0.27 & 0.27 \\
& 2. Scrambled input $s$ & 0.26 & \textbf{0.38} & - & 0.25 & 0.28 & \textbf{0.34} \\
& 3. Permuted $W$ columns & \underline{0.29} & 0.27 & - & \textbf{0.28} & \underline{0.31} & 0.30 \\
& 4. Heuristic Depth Mean & - & - & - & - & - & - \\
\addlinespace[4pt]
& \textbf{DFA (Zero-shot)} & \textbf{0.36} & \underline{0.33} & - & \underline{0.27} & 0.29 & 0.29 \\
& \textbf{DFA (Best)} & \textbf{0.36} & \underline{0.33} & - & \underline{0.27} & \textbf{0.31} & \underline{0.31} \\
\midrule
\multirow{6}{*}{NAP} & 1. Random $W$ (Lower Bound) & 0.25 & \underline{0.34} & - & 0.25 & \underline{0.29} & 0.25 \\
& 2. Scrambled input $s$ & \underline{0.25} & \textbf{0.41} & - & \underline{0.26} & 0.25 & \underline{0.34} \\
& 3. Permuted $W$ columns & 0.24 & 0.29 & - & 0.26 & 0.27 & 0.28 \\
& 4. Heuristic Depth Mean & - & - & - & - & - & - \\
\addlinespace[4pt]
& \textbf{DFA (Zero-shot)} & 0.23 & 0.28 & - & 0.26 & \textbf{0.40} & \textbf{0.38} \\
& \textbf{DFA (Best)} & \textbf{0.28} & 0.28 & - & \textbf{0.27} & \textbf{0.40} & \textbf{0.38} \\
\midrule
\multirow{6}{*}{NAP-IG-Activations} & 1. Random $W$ (Lower Bound) & 0.25 & \textbf{0.39} & - & 0.26 & \underline{0.28} & 0.25 \\
& 2. Scrambled input $s$ & \underline{0.25} & \underline{0.34} & - & 0.27 & 0.26 & \underline{0.29} \\
& 3. Permuted $W$ columns & 0.23 & 0.28 & - & \textbf{0.27} & 0.27 & 0.27 \\
& 4. Heuristic Depth Mean & - & - & - & - & - & - \\
\addlinespace[4pt]
& \textbf{DFA (Zero-shot)} & 0.23 & 0.30 & - & 0.26 & \textbf{0.35} & \textbf{0.34} \\
& \textbf{DFA (Best)} & \textbf{0.29} & 0.30 & - & \underline{0.27} & \textbf{0.35} & \textbf{0.34} \\
\bottomrule
\end{tabular}
}
\caption{Validation and Ablation Study for target model qwen2.5-3b (source qwen2.5-1.5b). We report the zero-shot faithfulness drop under various structural corruptions using CPR.}
\label{tab:ablations-summary-qwen2-5-1-5b-to-qwen2-5-3b}
\end{table*}

\FloatBarrier

\section{Generalization Across Model Scales and Families}
\begin{figure}[t]
  \centering
  \includegraphics[width=\linewidth]{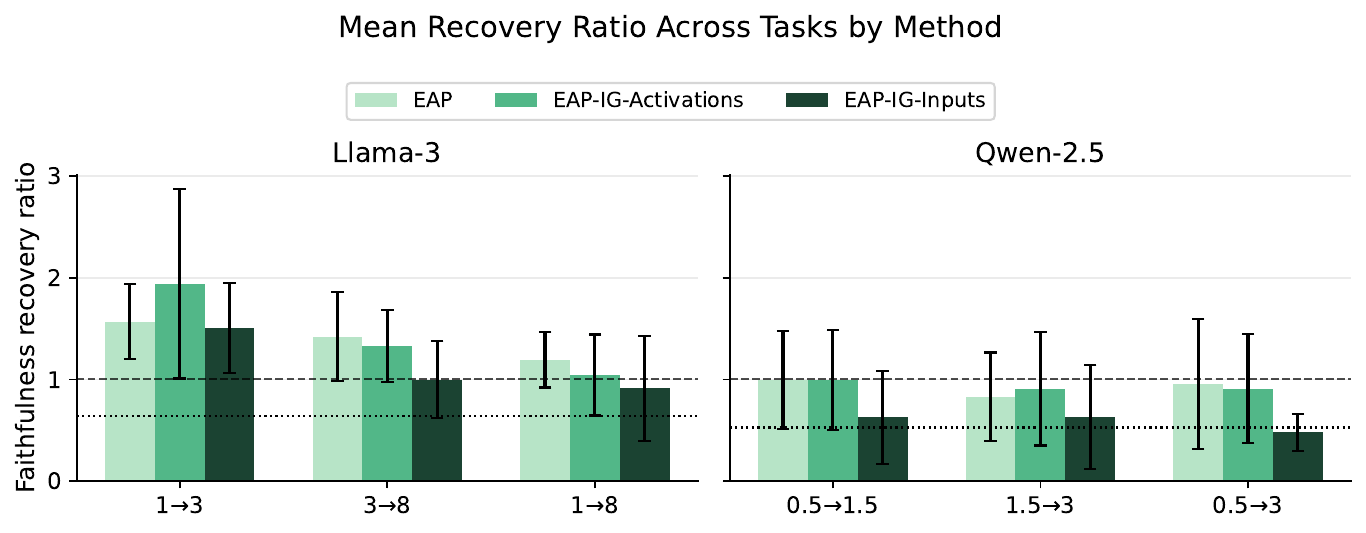}
    \caption{Mean faithfulness recovery ratio across tasks for each attribution method and source--target model pair, shown separately for the LLaMA-3 and Qwen-2.5 families. Recovery ratio is defined as $\text{Aligned(zero-shot)} / \text{Gold}$, where \textbf{Gold} is the direct target-model attribution result. Bars show the mean across the six tasks available for all compared settings, and error bars denote one standard deviation. The dashed line at $1.0$ marks parity with direct target attribution.}
  \label{fig:cross-model-recovery-mean-over-tasks-best}
\end{figure}

\begin{figure}[t]
  \centering
  \includegraphics[width=\linewidth]{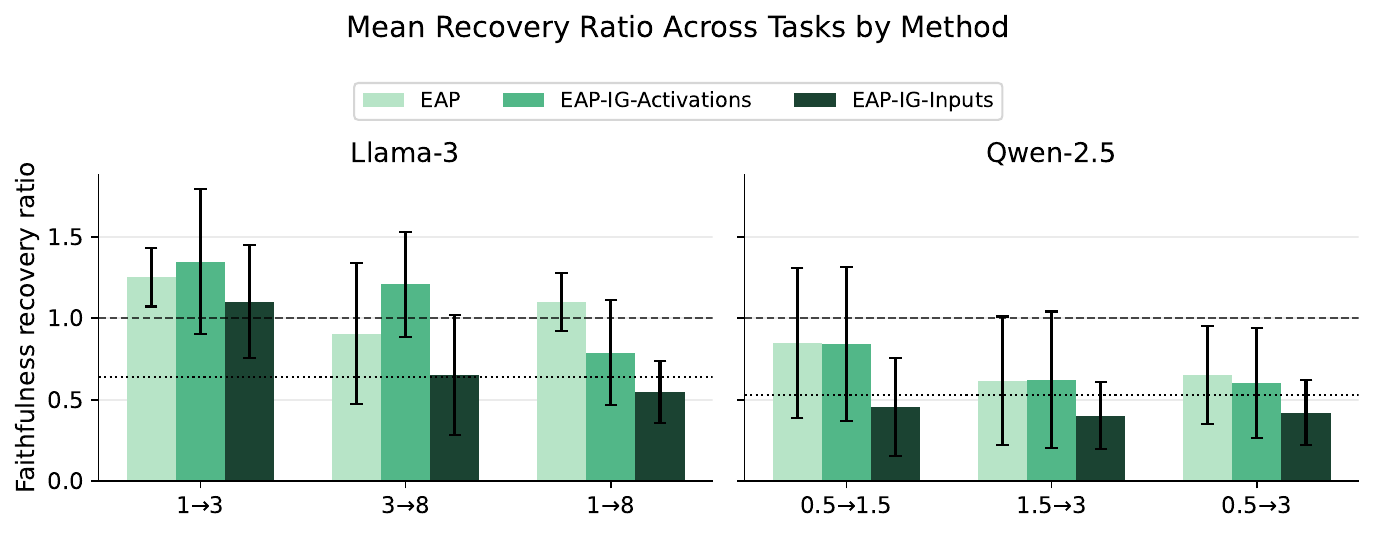}
    \caption{Mean faithfulness recovery ratio across tasks for each attribution method and source--target model pair, shown separately for the LLaMA-3 and Qwen-2.5 families. Recovery ratio is defined as $\text{Aligned(zero-shot)} / \text{Gold}$, where \textbf{Gold} is the direct target-model attribution result. Bars show the mean across the six tasks available for all compared settings, and error bars denote one standard deviation. The dashed line at $1.0$ marks parity with direct target attribution.}
  \label{fig:cross-model-recovery-mean-over-tasks-in-distribution}
\end{figure}

Tables \ref{tab:scaling-summary-mcqa-eap-eap-ig-activations}--\ref{tab:scaling-summary-arithmetic-subtraction} and Figure \ref{fig:cross-model-recovery-mean-over-tasks-best} and \ref{fig:cross-model-recovery-mean-over-tasks-in-distribution} summarize how recovery changes across model families, gaps, and tasks. The main text already says these appendix materials contain the broader scaling results.

\begin{table*}[!t]
\centering
\resizebox{\textwidth}{!}{%
\begin{tabular}{llcccc}
\toprule
\textbf{Method} & \textbf{Target Model Pair} & \textbf{Baseline} & \textbf{DFA (Best)} & \textbf{Gold} & \textbf{Faithfulness Recovery Ratio} \\
\midrule
\multirow{6}{*}{NAP-IG-Activations} & Llama-3 (1B $\to$ 3B) & 0.25 & 0.46 & 0.35 & 132.1\% \\
& Llama-3 (1B $\to$ 8B) & 0.25 & 0.40 & 0.60 & 66.6\% \\
& Llama-3 (3B $\to$ 8B) & 0.26 & 0.54 & 0.60 & 91.0\% \\
& Qwen-2.5 (0.5B $\to$ 1.5B) & 0.25 & 0.34 & 0.60 & 57.0\% \\
& Qwen-2.5 (0.5B $\to$ 3B) & 0.27 & 0.36 & 0.90 & 40.0\% \\
& Qwen-2.5 (1.5B $\to$ 3B) & 0.39 & 0.30 & 0.90 & 33.8\% \\
\midrule
\multirow{6}{*}{NAP} & Llama-3 (1B $\to$ 3B) & 0.25 & 0.49 & 0.35 & 139.7\% \\
& Llama-3 (1B $\to$ 8B) & 0.25 & 0.46 & 0.38 & 120.9\% \\
& Llama-3 (3B $\to$ 8B) & 0.25 & 0.32 & 0.38 & 85.3\% \\
& Qwen-2.5 (0.5B $\to$ 1.5B) & 0.25 & 0.28 & 0.56 & 49.1\% \\
& Qwen-2.5 (0.5B $\to$ 3B) & 0.27 & 0.37 & 0.82 & 45.8\% \\
& Qwen-2.5 (1.5B $\to$ 3B) & 0.34 & 0.28 & 0.82 & 34.8\% \\
\bottomrule
\end{tabular}%
}
\caption{Model scaling summary for MCQA. Best aligned performance for this task using CPR. Recovery\% $= \frac{\text{Aligned(best)}}{\text{Gold}}\times 100$.}
\label{tab:scaling-summary-mcqa-eap-eap-ig-activations}
\end{table*}

\begin{table*}[!t]
\centering
\resizebox{\textwidth}{!}{%
\begin{tabular}{llcccc}
\toprule
\textbf{Method} & \textbf{Target Model Pair} & \textbf{Baseline} & \textbf{DFA (Best)} & \textbf{Gold} & \textbf{Faithfulness Recovery Ratio} \\
\midrule
\multirow{6}{*}{NAP-IG-Inputs} & Llama-3 (1B $\to$ 3B) & 0.25 & 0.35 & 0.28 & 123.4\% \\
& Llama-3 (1B $\to$ 8B) & 0.25 & 0.34 & 0.63 & 53.5\% \\
& Llama-3 (3B $\to$ 8B) & 0.25 & 0.57 & 0.63 & 91.0\% \\
& Qwen-2.5 (0.5B $\to$ 1.5B) & 0.25 & 0.30 & 0.91 & 32.8\% \\
& Qwen-2.5 (0.5B $\to$ 3B) & 0.25 & 0.30 & 0.91 & 32.9\% \\
& Qwen-2.5 (1.5B $\to$ 3B) & 0.25 & 0.36 & 0.91 & 39.9\% \\
\midrule
\multirow{6}{*}{NAP-IG-Activations} & Llama-3 (1B $\to$ 3B) & 0.25 & 0.33 & 0.38 & 88.2\% \\
& Llama-3 (1B $\to$ 8B) & 0.25 & 0.43 & 0.35 & 124.7\% \\
& Llama-3 (3B $\to$ 8B) & 0.25 & 0.50 & 0.35 & 143.7\% \\
& Qwen-2.5 (0.5B $\to$ 1.5B) & 0.25 & 0.29 & 0.25 & 114.2\% \\
& Qwen-2.5 (0.5B $\to$ 3B) & 0.25 & 0.25 & 0.25 & 100.2\% \\
& Qwen-2.5 (1.5B $\to$ 3B) & 0.25 & 0.29 & 0.25 & 117.9\% \\
\midrule
\multirow{6}{*}{NAP} & Llama-3 (1B $\to$ 3B) & 0.25 & 0.32 & 0.25 & 128.8\% \\
& Llama-3 (1B $\to$ 8B) & 0.25 & 0.33 & 0.31 & 106.6\% \\
& Llama-3 (3B $\to$ 8B) & 0.25 & 0.36 & 0.31 & 116.4\% \\
& Qwen-2.5 (0.5B $\to$ 1.5B) & 0.25 & 0.28 & 0.25 & 113.7\% \\
& Qwen-2.5 (0.5B $\to$ 3B) & 0.25 & 0.32 & 0.25 & 128.4\% \\
& Qwen-2.5 (1.5B $\to$ 3B) & 0.25 & 0.28 & 0.25 & 110.5\% \\
\bottomrule
\end{tabular}%
}
\caption{Model scaling summary for IOI. Best aligned performance for this task using CPR. Recovery\% $= \frac{\text{Aligned(best)}}{\text{Gold}}\times 100$.}
\label{tab:scaling-summary-ioi}
\end{table*}

\begin{table*}[!t]
\centering
\resizebox{\textwidth}{!}{%
\begin{tabular}{llcccc}
\toprule
\textbf{Method} & \textbf{Target Model Pair} & \textbf{Baseline} & \textbf{DFA (Best)} & \textbf{Gold} & \textbf{Faithfulness Recovery Ratio} \\
\midrule
\multirow{6}{*}{NAP-IG-Inputs} & Llama-3 (1B $\to$ 3B) & 0.25 & 0.39 & 0.48 & 80.5\% \\
& Llama-3 (1B $\to$ 8B) & 0.25 & 0.38 & 0.94 & 41.0\% \\
& Llama-3 (3B $\to$ 8B) & 0.25 & 0.34 & 0.94 & 36.4\% \\
& Qwen-2.5 (0.5B $\to$ 1.5B) & 0.25 & 0.29 & 0.95 & 30.5\% \\
& Qwen-2.5 (0.5B $\to$ 3B) & 0.25 & 0.34 & 0.94 & 36.5\% \\
& Qwen-2.5 (1.5B $\to$ 3B) & 0.27 & 0.31 & 0.94 & 32.9\% \\
\midrule
\multirow{6}{*}{NAP-IG-Activations} & Llama-3 (1B $\to$ 3B) & 0.25 & 0.51 & 0.35 & 147.0\% \\
& Llama-3 (1B $\to$ 8B) & 0.25 & 0.40 & 0.63 & 64.1\% \\
& Llama-3 (3B $\to$ 8B) & 0.26 & 0.56 & 0.63 & 89.2\% \\
& Qwen-2.5 (0.5B $\to$ 1.5B) & 0.30 & 0.44 & 0.72 & 61.6\% \\
& Qwen-2.5 (0.5B $\to$ 3B) & 0.25 & 0.48 & 1.11 & 43.4\% \\
& Qwen-2.5 (1.5B $\to$ 3B) & 0.25 & 0.34 & 1.11 & 30.4\% \\
\midrule
\multirow{6}{*}{NAP} & Llama-3 (1B $\to$ 3B) & 0.28 & 0.52 & 0.34 & 151.7\% \\
& Llama-3 (1B $\to$ 8B) & 0.25 & 0.53 & 0.59 & 90.1\% \\
& Llama-3 (3B $\to$ 8B) & 0.25 & 0.29 & 0.59 & 49.3\% \\
& Qwen-2.5 (0.5B $\to$ 1.5B) & 0.27 & 0.36 & 0.67 & 53.5\% \\
& Qwen-2.5 (0.5B $\to$ 3B) & 0.25 & 0.42 & 1.10 & 37.8\% \\
& Qwen-2.5 (1.5B $\to$ 3B) & 0.25 & 0.38 & 1.10 & 34.6\% \\
\bottomrule
\end{tabular}%
}
\caption{Model scaling summary for ARC-C. Best aligned performance for this task using CPR. Recovery\% $= \frac{\text{Aligned(best)}}{\text{Gold}}\times 100$.}
\label{tab:scaling-summary-arc-challenge}
\end{table*}

\begin{table*}[!t]
\centering
\resizebox{\textwidth}{!}{%
\begin{tabular}{llcccc}
\toprule
\textbf{Method} & \textbf{Target Model Pair} & \textbf{Baseline} & \textbf{DFA (Best)} & \textbf{Gold} & \textbf{Faithfulness Recovery Ratio} \\
\midrule
\multirow{6}{*}{NAP-IG-Inputs} & Llama-3 (1B $\to$ 3B) & 0.25 & 0.54 & 0.52 & 104.1\% \\
& Llama-3 (1B $\to$ 8B) & 0.25 & 0.47 & 0.97 & 49.0\% \\
& Llama-3 (3B $\to$ 8B) & 0.25 & 0.34 & 0.97 & 34.7\% \\
& Qwen-2.5 (0.5B $\to$ 1.5B) & 0.25 & 0.29 & 0.91 & 32.2\% \\
& Qwen-2.5 (0.5B $\to$ 3B) & 0.25 & 0.35 & 1.06 & 32.6\% \\
& Qwen-2.5 (1.5B $\to$ 3B) & 0.27 & 0.31 & 1.06 & 29.6\% \\
\midrule
\multirow{6}{*}{NAP-IG-Activations} & Llama-3 (1B $\to$ 3B) & 0.26 & 0.41 & 0.43 & 97.3\% \\
& Llama-3 (1B $\to$ 8B) & 0.25 & 0.35 & 0.59 & 59.5\% \\
& Llama-3 (3B $\to$ 8B) & 0.26 & 0.56 & 0.59 & 95.6\% \\
& Qwen-2.5 (0.5B $\to$ 1.5B) & 0.28 & 0.42 & 0.76 & 55.4\% \\
& Qwen-2.5 (0.5B $\to$ 3B) & 0.25 & 0.34 & 0.90 & 38.5\% \\
& Qwen-2.5 (1.5B $\to$ 3B) & 0.28 & 0.35 & 0.90 & 38.7\% \\
\midrule
\multirow{6}{*}{NAP} & Llama-3 (1B $\to$ 3B) & 0.27 & 0.46 & 0.37 & 123.8\% \\
& Llama-3 (1B $\to$ 8B) & 0.25 & 0.49 & 0.57 & 87.3\% \\
& Llama-3 (3B $\to$ 8B) & 0.25 & 0.29 & 0.57 & 51.0\% \\
& Qwen-2.5 (0.5B $\to$ 1.5B) & 0.27 & 0.38 & 0.70 & 54.6\% \\
& Qwen-2.5 (0.5B $\to$ 3B) & 0.25 & 0.37 & 0.89 & 41.5\% \\
& Qwen-2.5 (1.5B $\to$ 3B) & 0.29 & 0.40 & 0.89 & 44.5\% \\
\bottomrule
\end{tabular}%
}
\caption{Model scaling summary for ARC-E. Best aligned performance for this task using CPR. Recovery\% $= \frac{\text{Aligned(best)}}{\text{Gold}}\times 100$.}
\label{tab:scaling-summary-arc-easy}
\end{table*}

\begin{table*}[!t]
\centering
\resizebox{\textwidth}{!}{%
\begin{tabular}{llcccc}
\toprule
\textbf{Method} & \textbf{Target Model Pair} & \textbf{Baseline} & \textbf{DFA (Best)} & \textbf{Gold} & \textbf{Faithfulness Recovery Ratio} \\
\midrule
\multirow{6}{*}{NAP-IG-Inputs} & Llama-3 (1B $\to$ 3B) & 0.25 & 0.51 & 0.26 & 196.5\% \\
& Llama-3 (1B $\to$ 8B) & 0.25 & 0.67 & 0.46 & 146.7\% \\
& Llama-3 (3B $\to$ 8B) & 0.25 & 0.45 & 0.46 & 98.9\% \\
& Qwen-2.5 (0.5B $\to$ 1.5B) &  &  &  &  \\
& Qwen-2.5 (0.5B $\to$ 3B) &  &  &  &  \\
& Qwen-2.5 (1.5B $\to$ 3B) &  &  &  &  \\
\midrule
\multirow{6}{*}{NAP-IG-Activations} & Llama-3 (1B $\to$ 3B) & 0.25 & 0.82 & 0.25 & 321.9\% \\
& Llama-3 (1B $\to$ 8B) & 0.25 & 0.41 & 0.33 & 121.5\% \\
& Llama-3 (3B $\to$ 8B) & 0.25 & 0.46 & 0.33 & 138.9\% \\
& Qwen-2.5 (0.5B $\to$ 1.5B) &  &  &  &  \\
& Qwen-2.5 (0.5B $\to$ 3B) &  &  &  &  \\
& Qwen-2.5 (1.5B $\to$ 3B) &  &  &  &  \\
\midrule
\multirow{6}{*}{NAP} & Llama-3 (1B $\to$ 3B) & 0.25 & 0.39 & 0.25 & 154.7\% \\
& Llama-3 (1B $\to$ 8B) & 0.25 & 0.40 & 0.25 & 161.5\% \\
& Llama-3 (3B $\to$ 8B) & 0.25 & 0.40 & 0.25 & 160.2\% \\
& Qwen-2.5 (0.5B $\to$ 1.5B) &  &  &  &  \\
& Qwen-2.5 (0.5B $\to$ 3B) &  &  &  &  \\
& Qwen-2.5 (1.5B $\to$ 3B) &  &  &  &  \\
\bottomrule
\end{tabular}%
}
\caption{Model scaling summary for Arith +. Best aligned performance for this task using CPR. Recovery\% $= \frac{\text{Aligned(best)}}{\text{Gold}}\times 100$.}
\label{tab:scaling-summary-arithmetic-addition}
\end{table*}

\begin{table*}[!t]
\centering
\resizebox{\textwidth}{!}{%
\begin{tabular}{llcccc}
\toprule
\textbf{Method} & \textbf{Target Model Pair} & \textbf{Baseline} & \textbf{DFA (Best)} & \textbf{Gold} & \textbf{Faithfulness Recovery Ratio} \\
\midrule
\multirow{6}{*}{NAP-IG-Inputs} & Llama-3 (1B $\to$ 3B) & 0.25 & 0.49 & 0.25 & 192.0\% \\
& Llama-3 (1B $\to$ 8B) & 0.25 & 0.39 & 0.37 & 106.3\% \\
& Llama-3 (3B $\to$ 8B) & 0.25 & 0.67 & 0.37 & 180.8\% \\
& Qwen-2.5 (0.5B $\to$ 1.5B) & 0.25 & 0.39 & 0.25 & 154.2\% \\
& Qwen-2.5 (0.5B $\to$ 3B) & 0.25 & 0.26 & 0.32 & 81.4\% \\
& Qwen-2.5 (1.5B $\to$ 3B) & 0.25 & 0.27 & 0.32 & 85.4\% \\
\midrule
\multirow{6}{*}{NAP-IG-Activations} & Llama-3 (1B $\to$ 3B) & 0.25 & 0.45 & 0.25 & 179.5\% \\
& Llama-3 (1B $\to$ 8B) & 0.25 & 0.28 & 0.31 & 89.9\% \\
& Llama-3 (3B $\to$ 8B) & 0.25 & 0.59 & 0.31 & 190.6\% \\
& Qwen-2.5 (0.5B $\to$ 1.5B) & 0.25 & 0.40 & 0.24 & 164.9\% \\
& Qwen-2.5 (0.5B $\to$ 3B) & 0.25 & 0.26 & 0.25 & 102.4\% \\
& Qwen-2.5 (1.5B $\to$ 3B) & 0.26 & 0.27 & 0.25 & 108.8\% \\
\midrule
\multirow{6}{*}{NAP} & Llama-3 (1B $\to$ 3B) & 0.25 & 0.26 & 0.25 & 104.7\% \\
& Llama-3 (1B $\to$ 8B) & 0.25 & 0.33 & 0.29 & 115.1\% \\
& Llama-3 (3B $\to$ 8B) & 0.25 & 0.40 & 0.29 & 139.8\% \\
& Qwen-2.5 (0.5B $\to$ 1.5B) & 0.25 & 0.40 & 0.24 & 162.6\% \\
& Qwen-2.5 (0.5B $\to$ 3B) & 0.25 & 0.26 & 0.25 & 102.7\% \\
& Qwen-2.5 (1.5B $\to$ 3B) & 0.25 & 0.27 & 0.25 & 109.2\% \\
\bottomrule
\end{tabular}%
}
\caption{Model scaling summary for Arith -. Best aligned performance for this task using CPR. Recovery\% $= \frac{\text{Aligned(best)}}{\text{Gold}}\times 100$.}
\label{tab:scaling-summary-arithmetic-subtraction}
\end{table*}

\FloatBarrier

\FloatBarrier

\section{Cross-Task Transfer Heatmap}
\begin{figure}[t]
  \centering
  \includegraphics[width=\linewidth]{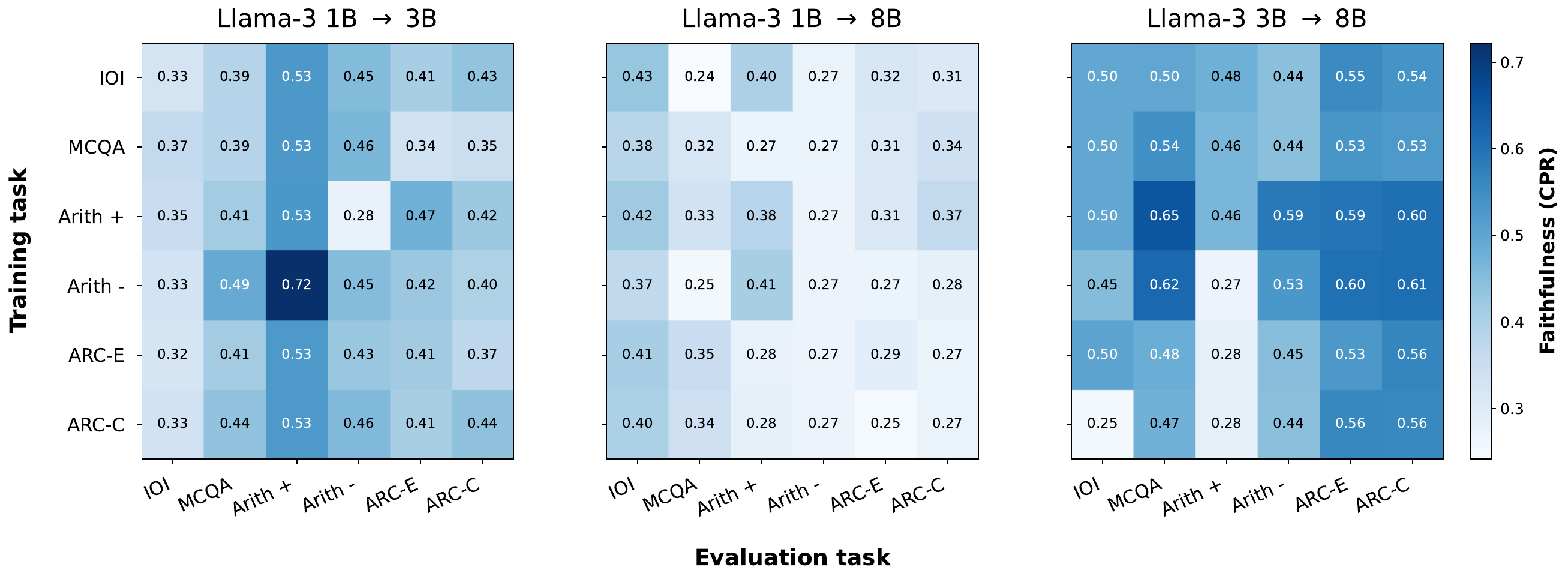}
    \caption{Cross-task transfer matrix for \textsc{EAP-IG-activations} on LLaMA-3, evaluated using \texttt{CPR}. Each row denotes the task used to train the alignment, and each column denotes the evaluation task. Entries report the faithfulness of the transferred circuit in the target model. Diagonal entries correspond to matched-task transfer, while off-diagonal entries measure cross-task generalization. The matrix is not symmetric, indicating that some tasks induce more reusable source--target correspondences than others.}
  \label{fig:cross-task-heatmap-llama-eap-ig-activations-llama}
\end{figure}

\begin{figure}[t]
  \centering
  \includegraphics[width=\linewidth]{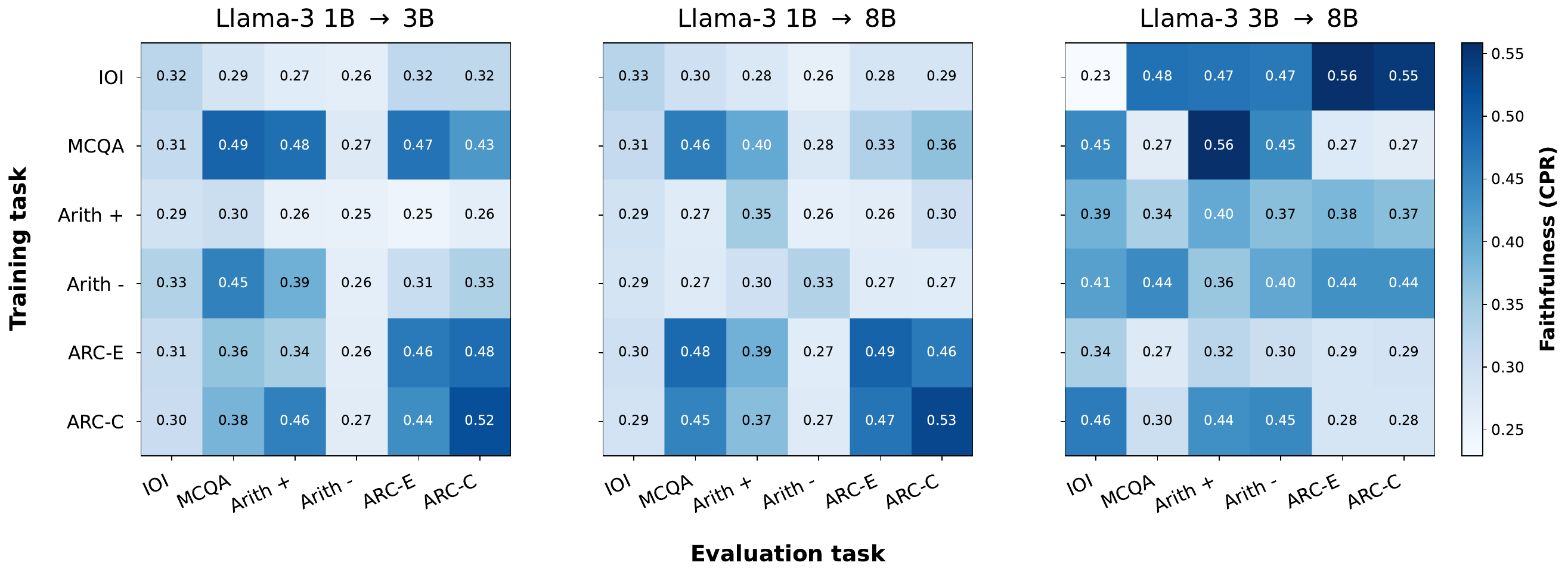}
    \caption{Cross-task transfer matrix for \textsc{EAP} on LLaMA-3, evaluated using \texttt{CPR}. Each row denotes the task used to train the alignment, and each column denotes the evaluation task. Entries report the faithfulness of the transferred circuit in the target model. Diagonal entries correspond to matched-task transfer, while off-diagonal entries measure cross-task generalization. The matrix is not symmetric, indicating that some tasks induce more reusable source--target correspondences than others.}
  \label{fig:cross-task-heatmap-llama-eap-llama}
\end{figure}

\begin{figure}[t]
  \centering
  \includegraphics[width=\linewidth]{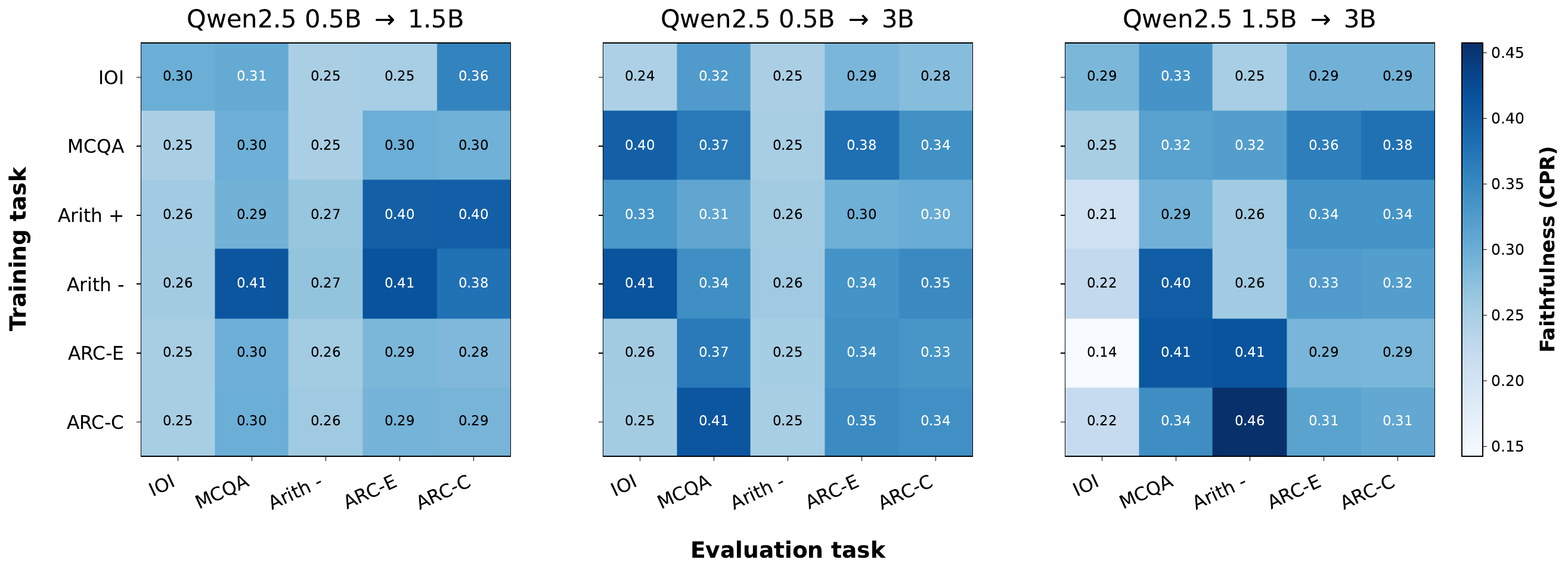}
    \caption{Cross-task transfer matrix for \textsc{EAP-IG-inputs} on Qwen-2.5, evaluated using \texttt{CPR}. Each row denotes the task used to train the alignment, and each column denotes the evaluation task. Entries report the faithfulness of the transferred circuit in the target model. Diagonal entries correspond to matched-task transfer, while off-diagonal entries measure cross-task generalization. The matrix is not symmetric, indicating that some tasks induce more reusable source--target correspondences than others.}
  \label{fig:cross-task-heatmap-llama-eap-ig-inputs-qwen}
\end{figure}

\begin{figure}[t]
  \centering
  \includegraphics[width=\linewidth]{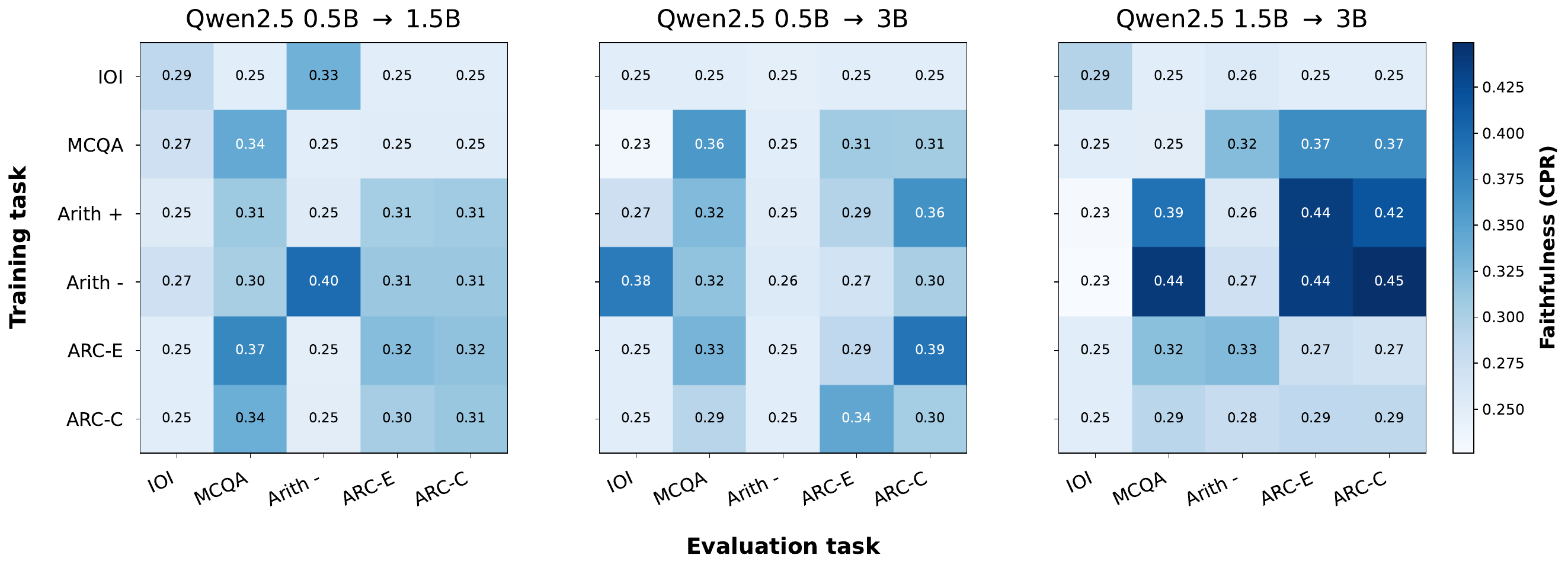}
    \caption{Cross-task transfer matrix for \textsc{EAP-IG-activations} on Qwen-2.5, evaluated using \texttt{CPR}. Each row denotes the task used to train the alignment, and each column denotes the evaluation task. Entries report the faithfulness of the transferred circuit in the target model. Diagonal entries correspond to matched-task transfer, while off-diagonal entries measure cross-task generalization. The matrix is not symmetric, indicating that some tasks induce more reusable source--target correspondences than others.}
  \label{fig:cross-task-heatmap-llama-eap-ig-activations-qwen}
\end{figure}

\begin{figure}[t]
  \centering
  \includegraphics[width=\linewidth]{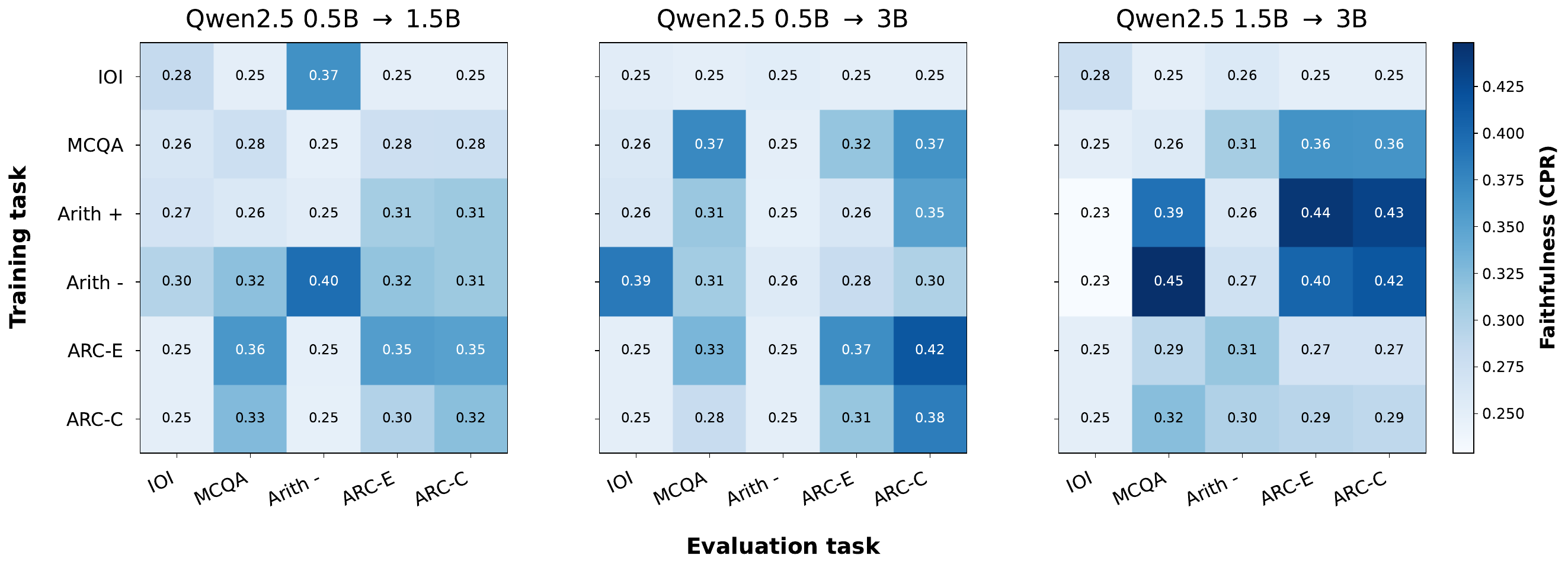}
    \caption{Cross-task transfer matrix for \textsc{EAP} on Qwen-2.5, evaluated using \texttt{CPR}. Each row denotes the task used to train the alignment, and each column denotes the evaluation task. Entries report the faithfulness of the transferred circuit in the target model. Diagonal entries correspond to matched-task transfer, while off-diagonal entries measure cross-task generalization. The matrix is not symmetric, indicating that some tasks induce more reusable source--target correspondences than others.}
  \label{fig:cross-task-heatmap-llama-eap-qwen}
\end{figure}

We provide additional task-to-task transfer matrices for attribution methods and model pairs not shown in the main text. Figures \ref{fig:cross-task-heatmap-llama-eap-ig-activations-llama}--\ref{fig:cross-task-heatmap-llama-eap-qwen} complement Figure \ref{fig:cross-task-heatmap-llama-eap-ig-inputs-llama} by showing how the learned alignment transfers across training and evaluation tasks beyond the Llama-3 1B→3B NAP-IG-Inputs setting. These heatmaps help clarify which transfer patterns are stable across methods and which are specific to particular source–target pairs.

\FloatBarrier

\section{Alignment beyond circuit analysis}

DFA uses attribution scores from a smaller model to predict which components matter in a larger one, where repeating the analysis for each task is costly. A similar motivation underlies efforts to predict large-model generalization
error from small-scale experiments \citep{rosenfeld2020iclr-constructive}.

Alignment can also help when direct supervision is unavailable. AMSAL \citep{shao-etal-2022-SAL} aligns representations with unpaired protected-attribute values \citep{shao2023erasure} before applying spectral attribute removal . H-SAL \citep{shao2026hsal} uses free-form author profiles to identify directions to erase, while DAFair \citep{iskander-etal-2024-leveraging} uses textual group prototypes to guide debiasing during fine-tuning. IMSAE likewise seeks to identify how attribute-related directions align across languages within a single multilingual model. It estimates joint
bias subspaces over subsets of languages and composes the resulting projections into a global debiasing transformation \citep{shao-etal-2025-iterative}.

\end{document}